**A Review of Learning-Based Motion Planning: Toward a Data-Driven Optimal Control Approach**

**Jia Hu, Ph.D**
ZhongTe Distinguished Chair in Cooperative Automation, Professor
College of Transportation
Key Laboratory of Road and Traffic Engineering of the Ministry of Education
Institute for Advanced Study
Tongji University
4800 Cao'an Road, Shanghai, P.R.China
Email: hujia@tongji.edu.cn

**Yang Chang**
College of Transportation
Key Laboratory of Road and Traffic Engineering of the Ministry of Education
Tongji University
4800 Cao'an Road, Shanghai, P.R.China
Email: Neymar66@tongji.edu.cn

**Haoran Wang, Ph.D, Corresponding Author**
Associate Professor
College of Transportation
Key Laboratory of Road and Traffic Engineering of the Ministry of Education
Tongji University
4800 Cao'an Road, Shanghai, P.R.China
Email: wang_haoran@tongji.edu.cn

**Abstract**

Motion planning for autonomous driving (AD) faces a critical trade-off. While traditional rule-based pipelines offer verifiable safety and interpretability, they often fail to generalize in complex scenarios. Conversely, emerging learning-based methods—including imitation learning (IL), reinforcement learning (RL), and generative AI—offer greater adaptability but are often constrained by opacity and safety risks. Existing surveys typically analyze these AI methods in isolation, overlooking the potential of integrating them with rigorous control frameworks. To bridge this gap, this paper presents the first systematic review of the Data-Driven Optimal Control (DDOC) paradigm, explicitly examining how it synergizes the theoretical guarantees of optimal control with the adaptive capabilities of modern machine learning. Building on this framework, we propose the first roadmap for DDOC-based motion planning, structuring its implementation into three critical dimensions: customization, dynamics adaptation, and self-tuning. Finally, to close the remaining reality gap, we identify four future research directions, thereby accelerating the transition to trustworthy and human-like autonomous driving.

## 1. Introduction

Autonomous Driving (AD) stands as a pivotal technology for future Intelligent Transportation Systems (ITS). It has garnered significant attention and investment from governments, academia, and industry. Projections by Roland Berger suggest that over 70% of vehicles worldwide will be equipped with automated applications, creating a market exceeding $4 trillion (Limited, 2023). The advancement of AD promises to revolutionize daily life by alleviating driving burdens and ultimately achieving full autonomy.

However, significant obstacles remain. Currently, the commercialization of high-level automation faces substantial hurdles. In real-world environments, AD systems must adapt to complex, long-tail scenarios. Preparing for all these scenarios has been a great challenge in actual implementations. Furthermore, unpredictable behaviors by Autonomous Vehicles (AVs) can disrupt surrounding traffic. For example, AVs are often impeded by pedestrians, cyclists, and Human-Driven Vehicles (HVs) at intersections, potentially impacting normal traffic flow. It would impact the regular operation of traffic. Therefore, there is a growing need for the enhancement of AD's motion planning capability.

In recent years, Artificial Intelligence (AI) has emerged as a focal point of research and is regarded as a primary avenue for advancing AD. AI has already been widely deployed in perception and prediction, where many solutions have reached commercial maturity. Consequently, learning-based motion planning has become a subject of extensive study. They commonly incorporate AI techniques such as Reinforcement Learning (RL) (Li et al., 2020; Naveed et al., 2021; Panov et al., 2018), and Imitation Learning (IL) (Kocić et al., 2019; Lee and Ha, 2020; Song et al., 2019) into motion planning problems. Nevertheless, learning-based motion planning still faces major challenges in safety guarantees, training efficiency, and reliable performance. AI models are often criticized as "black boxes" that lack interpretability. To ensure driving safety in the field implementation of AD, learning-based motion planning methods mostly focus on the upper-level problems, including path planning (Panov et al., 2018) and maneuver planning (Naveed et al., 2021). Conventional technologies, such as the rule-based method, sampling method, and optimal control-based method, are adopted in the lower-level planning as a safety layer. Despite their current limitations in commercial applications, learning-based approaches are expected to address critical AD challenges, including human-like driving, interaction, and adaptation to long-tail scenarios.

Therefore, identifying a feasible roadmap utilizing AI is imperative to accelerate high-level automation. A comprehensive literature review is essential to delineate this path. As illustrated in **Table 1,** existing reviews predominantly analyze specific learning-based methods in isolation or provide broad overviews. For example, many works focus exclusively on RL (Kiran et al., 2021), IL (Le Mero et al., 2022), or the emerging role of Large Language Models (LLMs) (Yang et al., 2023; Zhu et al., 2025). While comprehensive surveys like those by Teng et al. (Teng et al., 2023) and Chen et al. (Chen et al., 2024c) catalog the broader field, they fail to propose a unifying roadmap that addresses the critical conflict between transparent, rigid pipelines and adaptive but opaque learning-based systems.

**Table 1** A comparative analysis of this paper against existing surveys in AD motion planning

| Survey | Scope | Pipeline | IL | RL | GAI/LLM | Data-Driven Optimal Control | Future roadmap |
|---|---|---|---|---|---|---|---|
| Kiran et al. (2021) | Motion planning | △ | ✓ | ✓ | × | × | × |
| Le Mero et al. (2022) | End-to-end | △ | ✓ | △ | × | △ | × |
| Teng et al. (2023) | Motion planning | ✓ | ✓ | ✓ | × | × | × |
| Chib and Singh (2023) | End-to-end | △ | ✓ | ✓ | △ | × | × |
| Cui et al. (2024) | End-to-end | × | × | × | ✓ | × | × |
| Yang et al. (2023) | Decision-making & planning | ✓ | △ | △ | ✓ | △ | × |
| Chen et al. | End-to-end | △ | ✓ | ✓ | ✓ | × | × |
| (2024c) | | | | | | | |
| Jiang et al. (2025) | End-to-end | ✓ | △ | △ | ✓ | △ | × |
| Zhu et al. (2025) | Motion planning | ✓ | × | × | ✓ | × | × |
| Ours | Motion planning | ✓ | ✓ | ✓ | ✓ | ✓ | The adoption of DDOC in AD motion planning |

Note: ✓ indicates primary focus/extensive coverage; △ indicates partial involvement; × indicates a brief mention or absence.

In contrast, this work presents the first joint review of IL, RL, and Generative AI through the lens of a unified Data-Driven Optimal Control (DDOC) paradigm. We explicitly examine how this integrated approach enables three critical next-generation capabilities: customization, dynamics adaptation, and self-tuning. By doing so, we aim to provide a feasible roadmap that resolves the impasse between the transparency of traditional pipelines and the adaptability of modern learning-based systems.

This paper reviews learning-based motion planning methods and offers the following contributions:

- It provides the first systematic review of DDOC research in motion planning and proposes a roadmap for DDOC-based AD motion planning.
- It systematically reviews key technologies in AD motion planning, spanning the pipeline methods to mainstream AI methods such as IL, RL, and Generative AI.
- It elaborates on the feasibility of this roadmap and analyzes the implementation of DDOC across three dimensions: customization, dynamics adaptation, and self-tuning.
- It outlines the evolutionary roadmap for the next-generation DDOC system, proposing four deeply integrated directions—Generative World Model-based prediction, Hybrid Learning for intent alignment, RL-based Meta-Control, and Certified Safety Verification—to bridge the reality gap and achieve trustworthy autonomy.

This review adopts a systematic literature search methodology. A comprehensive search was conducted across multiple electronic databases using the search strategy summarized in **Table 2**, covering keywords related to AD and motion planning, learning-based methods, and on-road driving scenarios. The search encompassed publications from January 2020 to August 2025, providing an overview of the research landscape over the past five years. Only English-language studies focusing on learning-based or pipeline-based motion planning for on-road AD were considered. Studies that addressed only perception, prediction, or low-level lane-keeping control without decision-making or interaction modeling were excluded. In total, 13,583 papers were identified initially. After title–abstract screening and full-text review, 641 papers were selected for the final analysis. The screening flow diagram and keyword co-occurrence network are presented in **Fig. 1** and **Fig. 2**, respectively.

**Table 2** Database and search terms

| Search strategy | Details |
|---|---|
| Databases and search engines | IEEE Xplore, Web of Science, Scopus, arXiv, and Wiley Online Library |
| Search terms | Key concepts and their synonyms (autonomous driving [Title/Abstract/Keywords] OR automated driving [Title/Abstract/Keywords] OR motion planning [Title/Abstract/Keywords])<br>Methodologies (learning-based [Title/Abstract/Keywords] OR reinforcement learning [Title/Abstract/Keywords] OR imitation learning [Title/Abstract/Keywords] OR deep learning [Title/Abstract/Keywords] OR pipeline methods [Title/Abstract/Keywords] OR large language models [Title/Abstract/Keywords] OR optimal control [Title/Abstract/Keywords]) |

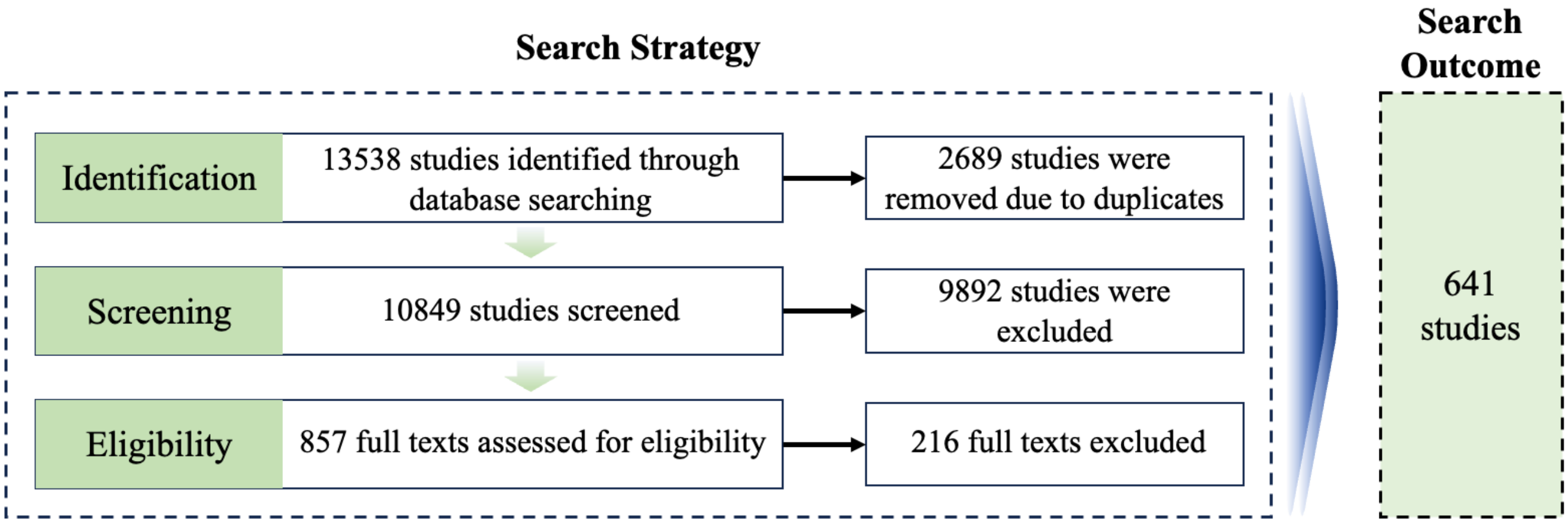

**Fig. 1.** Literature search processes

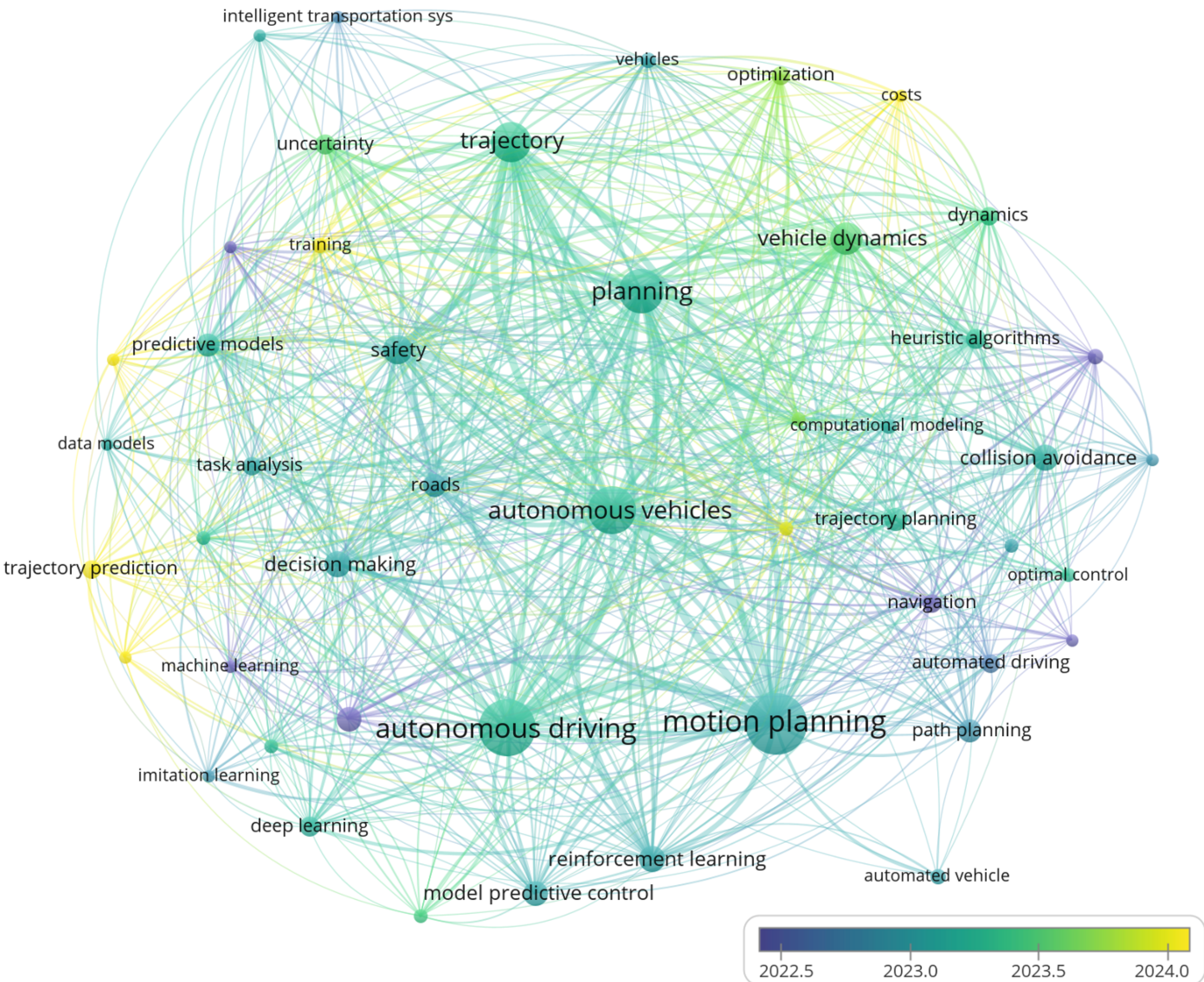

**Fig. 2.** Six-year (2020–2025) keyworks co-occurrence network for "autonomous driving motion planning"

The remainder of this paper is organized as follows: Section 2 deconstructs the canonical motion planning pipeline, establishing the functional expectations and fundamental challenges—such as generalizability, safety, and real-world deployment—that modern systems must address. Section 3 critically reviews the current learning-based landscape, specifically examining the progress and inherent limitations of IL, RL, and Generative AI. Section 4 introduces the central thesis of this work: the DDOC paradigm. We elaborate on its feasibility and analyze its implementation across three key dimensions. Finally, Section 5 outlines future research directions, followed by concluding remarks in Section 6.

## 2. Motion Planning Pipeline and Functional Expectations

Motion planning constitutes the cognitive core of the AD system, directly governing vehicle safety, efficiency, and ride comfort. Leveraging data from perception and localization modules, motion planning is required to compute a sequence of trajectory waypoints alongside the corresponding acceleration and steering commands for execution. However, a successful motion planning architecture must not only fulfill these fundamental objectives but also exhibit robust adaptability to the complex and stochastic nature of real-world traffic environments.

To systematically deconstruct the current state of the art, we first introduce the canonical motion planning pipeline, which modularizes the problem into route planning, behavior planning, and trajectory planning. This modular framework has served as the foundation for traditional systems. Yet, the pursuit of higher-level intelligence and adaptability has given rise to a spectrum of learning-based paradigms.

This paper will, therefore, provide a comprehensive review of the primary methodologies employed to solve the motion planning problem, including pipeline methods, IL, RL, and the more recent generative AI approaches. Our analysis will be framed by the critical challenges that persist in the field: generalizability, safety, interpretability, real-world deployment, interactive driving, and customization. By examining how each methodology addresses or falls short in the face of these challenges, we establish the necessary context for introducing a more integrated and powerful alternative: the data-driven optimal control approach, which forms the central thesis of this paper. This structured overview will illuminate the limitations of existing solutions and motivate the subsequent discussion on a more holistic and data-centric future for motion planning. To facilitate reading and ensure clarity throughout the discussion, **Table 3** provides a comprehensive list of the acronyms and corresponding definitions used in this paper.

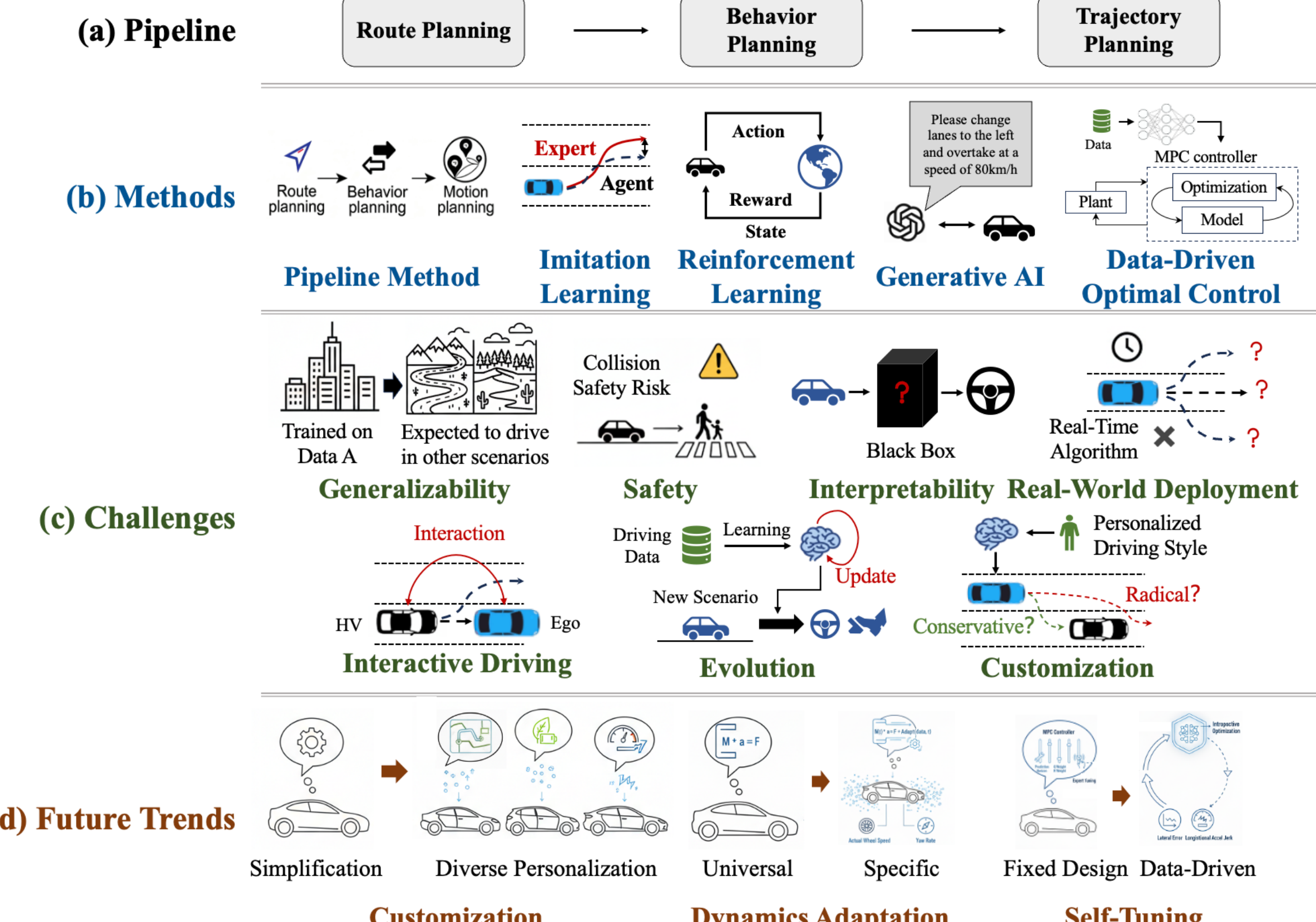

**Fig. 3.** An overview of the motion planning landscape, detailing (a) motion planning pipeline, (b)

competing methodologies, (c) key research challenges, and (d) future trends.

**Table 3** List of Acronyms and Definitions

| Acronym | Definition / Full name | Acronym | Definition / Full name |
|---|---|---|---|
| A3C | Asynchronous Advantage Actor-Critic | IL | Imitation Learning |
| ACC | Adaptive Cruise Control | IRL | Inverse Reinforcement Learning |
| AD | Autonomous Driving | LTI | Linear Time-Invariant |
| AI | Artificial Intelligence | ITS | Intelligent Transportation System |
| BC | Behavioral Cloning | LBC | Learning by Cheating |
| BEV | Bird's-Eye-View | LBMPC | Learning-Based Model Predictive Control |
| CIC | Conditional Imitation Co-learning | LLMs | Large Language Models |
| CIL | Conditional Imitation Learning | MDPs | Markov Decision Processes |
| CIRL | Controllable Imitation Reinforcement Learning | MLLMs | Multimodal Large Language Models |
| CNN | Convolutional Neural Network | MPC | Model Predictive Control |
| CoT | Chain-of-Thought | NNs | Neural Networks |
| DAgger | Dataset Aggregation | OTA | Over-The-Air |
| DDC | Data-Driven Control | ODDs | Operational Design Domains |
| DDPC | Data-Driven Predictive Control | PE | Persistent Excitation |
| DDPG | Deep Deterministic Policy Gradient | PER | Prioritized Experience Replay |
| DeePC | Data-enabled Predictive Control | PHEVs | Plug-in Hybrid Electric Vehicles |
| DNNs | Deep Neural Networks | PPO | Proximal Policy Optimization |
| DPL | Direct Policy Learning | RL | Reinforcement Learning |
| DQN | Deep Q-Networks | RNN | Recurrent Neural Network |
| DRL | Deep Reinforcement Learning | SL | Supervised Learning |
| GAIL | Generative Adversarial Imitation Learning | TD | Temporal Difference |
| GANs | Generative Adversarial Networks | TD3 | Twin Delayed Deep Deterministic Policy Gradient |
| GP | Gaussian Processes | VAEs | Variational Autoencoders |
| HCPI-RL | High-Confidence Policy Improvement Reinforcement Learning | VLAs | Vision-Language-Action Models |
| HRL | Hierarchical Reinforcement Learning | VLMs | Vision-Language Models |

## 2.1. Motion Planning Pipeline

Currently, a widely adopted pipeline of motion planning consists of three modules: route planning, behavior planning, and trajectory planning. Route planning generates a global path from the start point to the end point, in order to accomplish the passenger's travel task. Behavior planning makes decisions on driving behavior, such as following, lane-changing, and yielding. It operates in a microscopic driving environment that contains other traffic participants. Trajectory planning calculates real-time and short-term driving trajectories to accomplish planned behaviors. The planned trajectory shall consider the interaction with other traffic participants, avoiding collisions and mobility reductions due to excessive yielding. In this section, we discuss current technologies for the three tasks of motion planning.

### *2.1.1. Route Planning*

Route planning is fundamentally a static problem, as it operates independently of real-time interactions with other traffic participants (Igneczi et al., 2024). Route planning technologies have been well developed and commercialized. Dijkstra algorithm, A-star algorithm (Zhang et al., 2023b), and D-star algorithm are representative methods for route planning. They aim at finding the shortest path to the destination. Query time, solving efficiency, preprocessing complexity, and memory occupancy are the main focuses in the commercialization. The key challenge in implementation is the input for route planning, that is, the detection of road marking and vehicles' locating in this relative framework. This is quite important since AD is now experiencing progress without relying on HD maps. Furthermore, AD route planning is expected to find a better global path that is time-saving and energy-saving. It is another problem of cooperative automation, facilitated by vehicle-road-cloud cooperation. This topic is out of the scope of this research. Detailed discussion could be found in (Wang et al., 2024; Wang et al., 2022c).

### *2.1.2. Behavior Planning*

Behavior planning serves as the bridge between global routing and local trajectory. Background vehicles' intentions need to be predicted before planning (Lucente et al., 2023). Game theory is usually adopted to generate behaviors in an interactive driving environment (Li et al., 2024a). Considering real-time background traffic, behavior planning provides behavioral decisions that must be feasible for local trajectory planning. Hence, a perfect solution is to integrate behavior planning and trajectory planning (Wang et al., 2023). In this problem, the trajectory should be directly planned within the feasible driving region. It would be a complex optimization problem with low computational efficiency. Therefore, in the actual implementation, meta rules are widely adopted to simplify lower-level trajectory planning.

### *2.1.3. Trajectory Planning*

Trajectory planning is the core module of motion planning. It generates desired trajectories (position and time) for tracking (Liu et al., 2023a) or motion commands (acceleration and steering) to vehicle local control (Chen et al., 2024a). Nominally, trajectory planning is formulated as an optimal control problem, comprising the objective function, vehicle dynamics model, and collision avoidance (Li et al., 2023a). A trajectory prediction model is generally needed as the input of trajectory planning (An et al., 2024; Min et al., 2024; Peng et al., 2024). The optimal solution is the planned trajectories, presented as control and state profiles. The optimal trajectory planning problem is typically a time-continuous form (Wang et al., 2022b), which could be discretized for solving as shown in Eq. (1):

$$\begin{aligned} &\min_{\boldsymbol{u}_i} \sum_{\mathrm{i}=1}^{\mathrm{N}-1} J_i(\boldsymbol{x_i}, \boldsymbol{u}_i) + J_N(\boldsymbol{x_N}) \\ &\text{s.t. } f(\boldsymbol{x_i}, \boldsymbol{u}_i) = \boldsymbol{x}_{i+1} \\ &\quad \underline{\boldsymbol{x}} \le \boldsymbol{x_i} \le \overline{\boldsymbol{x}} \\ &\quad \underline{\boldsymbol{u}} \le \boldsymbol{u_i} \le \overline{\boldsymbol{u}} \\ &\quad \boldsymbol{x_i} \in \Omega_{free} \\ &\quad i = 1, \ldots, N-1 \end{aligned} \tag{1}$$

where $i$ denotes the index of control step; $N$ represents the prediction horizon; $\boldsymbol{x}_i$ denotes the vehicle state vector at time step $i$, belonging to the state space $\mathbb{R}^{n_x}$; $\boldsymbol{u}_i$ denotes the vehicle control vector at time step $i$, belonging to the control space $\mathbb{R}^{n_u}$; $J$ denotes the cost function; $f$ denotes the vehicle dynamics function; $\underline{\boldsymbol{x}}$ and $\overline{\boldsymbol{x}}$ denote the lower-bound and higher-bound of state vector, respectively; $\underline{\boldsymbol{u}}$ and $\overline{\boldsymbol{u}}$ denote the lower-bound and higher-bound of control vector, respectively; $\Omega_{free}$ denote the free state space excluding the collision space. The vehicle dynamics model is derived from the differential of the state vector in the continuous domain. The continuous form $\dot{\boldsymbol{x}}$ could be discretized based on methods like Euler.

## 2.2. Expectations for Motion Planning

Beyond fulfilling the basic function, any advanced motion planning system must first address a set of fundamental requirements to be considered robust, reliable, and effective in the real world. These foundational challenges include the ability to work properly on actual vehicles, generalize in countless scenarios, and ensure safety under all conditions. Failing to meet these basic criteria can lead to significant safety risks, especially in complex or unforeseen situations. Building upon this foundation, the next frontier for motion planning is to achieve a higher level of intelligence that emulates human-like driving capabilities. This advanced goal encompasses sophisticated interaction, continuous self-evolution, and deep personalization, which together define a truly mature and user-accepted AD experience.

### *2.2.1. Generalization and Adaptability*

A primary challenge for motion planning is generalizing its performance from seen training data to the infinite variations of the real world. Autonomous vehicles must adapt to a vast spectrum of complex driving scenarios, many of which are impossible to fully cover during development and testing. Conventional rule-based strategies, for instance, cannot be prepared for all possible situations (Hallgarten, 2025), creating significant safety risks in long-tail scenarios where unusual events occur. The learning-based approach has demonstrated promising results in addressing interactive and long tail scene problems (Lan et al., 2024; Zheng et al., 2024b). However, real-world data is often non-independent and identically distributed (non-i.i.d.), incomplete, and heterogeneous. The basic assumptions of statistics are no longer applicable, resulting in a lack of theoretical basis for the generalization ability of neural network models (Chu et al., 2024). Therefore, a core requirement is to develop motion planners that are not brittle but can robustly and safely handle novel traffic patterns, environmental conditions, and interactions.

*2.2.2. Safety and Interpretability*

Safety remains the paramount, non-negotiable requirement for AD, with system interpretability serving as a critical determinant. However, the increasing reliance on AI techniques presents a formidable challenge in this domain due to their "black-box" nature. AI-based planners, particularly end-to-end systems, are often criticized for their lack of interpretability (Teng et al., 2023). This opacity makes formal safety verification and debugging extremely challenging, as it is hard to guarantee the system's behavior in edge cases. In fact, to mitigate this risk in current field implementations, learning-based methods are often confined to higher-level tasks like maneuver planning, while conventional, more predictable methods are used as a lower-level safeguard (Chen and Aksun-Guvenc, 2025; Klimke et al., 2023). Overcoming this challenge is essential for building trustworthy systems and gaining public acceptance.

*2.2.3. Real-World Deployment*

Transitioning a motion planning algorithm from simulation to a production vehicle introduces a host of practical challenges. Firstly, there is the "sim-to-real" gap; a policy that performs perfectly in a simulator may fail in the real world due to subtle differences in sensor noise, vehicle dynamics, and the behavior of other road users. Secondly, production vehicles have limited onboard computational power, which constrains the complexity of the algorithms that can be run in real-time. This necessitates high training and execution efficiency. Effectively addressing these deployment challenges is crucial for the successful commercialization of high-level automation.

*2.2.4. Interactive Driving*

In real-world environments, human drivers constantly interact with other road users, including vehicles, pedestrians, and riders. For example, when we want to cut in, we may steer the wheel and make an attempt to see whether the other vehicle will yield. The logic behind the interactive driving is that the AV should be capable of understanding other agents' reacting rules, rather than predicting other agents' behaviors. Similar capability has emerged in some state-of-the-art AD applications, such as socially interactive decision-making frameworks (Schwarting et al., 2019), game-theoretic planning approaches (Tian et al., 2022; Yuan and Shan, 2024), and interaction-aware trajectory planning(Huang et al., 2023). However, these products still lack an understanding of other agents' reactions in the trajectory planning.

*2.2.5. Evolution*

AD is expected to adapt to various driving scenarios that may not be involved in pre-training. Hence, evolution capability is required to enable the accumulation of experience during implementation. Furthermore, the evolution of AD cannot be fulfilled by Over-The-Air (OTA) update, which is widely adopted in current commercial applications, since OTA cannot timely adapt to a fast-changing driving environment. This demand for real-time adaptation introduces significant challenges. True online evolution requires that any learning process is not only efficient but also verifiably safe. The training and inference must be highly efficient to operate within the constraints of onboard computation resources without impeding other critical vehicle functions. Moreover, the evolutionary process must be strictly monotonic. It enables the evolutionary policy to avoid deteriorating with more experience to avoid safety risks caused by the local optimum.

*2.2.6. Customization*

User experience has emerged as a distinguishing objective within the increasingly competitive AD market. It requires an adaptive AD to users' personalized preferences, also known as user customization. However, Current approaches to personalization are often limited to a few discrete, pre-defined settings, such as allowing a driver to choose between several following distances in Adaptive Cruise Control (ACC) or selecting a driving mode like "Sport", "Normal", or "Eco". The challenge lies in transcending discrete adjustments to achieve continuous, dynamic personalization. The goal is to create a system that continuously learns and adapts to an individual's unique driving style. For example, the vehicle should be able to infer a user's context-dependent preferences, such as prioritizing efficiency during weekday commutes while favoring comfort during weekend leisure trips. Achieving this level of nuanced adaptation is intrinsically linked to the system's evolution capability, with human-machine shared control offering a promising avenue for implementation(Li et al., 2023b; Song et al., 2024). Furthermore, they are both expected to be on board, no longer relying on OTA.

*2.2.7. Comparative Analysis of Methodologies*

As mentioned above, an ideal motion planning system for AD must simultaneously achieve several critical goals: it must guarantee safety and explainability, exhibit robust generalization and adaptability to novel scenarios, and maintain computational efficiency for real-world deployment. On this basis, achieving human-like driving is a more advanced goal. However, existing methodologies often force a compromise among these objectives.

Pipeline methods, while offering high levels of safety and verifiability through their structured, rule-based nature, are inherently brittle. They struggle to generalize to scenarios not explicitly programmed, making them inadequate for the complexity and unpredictability of the real world. Conversely, end-to-end learning-based approaches—such as IL, RL, and generative AI—demonstrate superior adaptability by learning complex behaviors directly from data. Yet, their "black box" nature creates a significant barrier to deployment, as it prevents formal safety verification and makes them prone to unpredictable failures, such as covariate shift or hallucinations, fundamentally compromising their trustworthiness in safety-critical applications.

This creates a crucial dilemma between the transparent but rigid pipeline and the adaptive but opaque learning models. Data-driven optimal control emerges as a promising paradigm to resolve this conflict. By integrating the principles of classical control theory with modern machine learning, this approach aims to achieve the best of both worlds. DDOC leverages a structured optimal control framework that preserves explainability and provides a mechanism for enforcing explicit safety constraints. Within this framework, it uses data-driven techniques to learn and adapt the system's internal models and parameters. This synergistic design directly addresses the brittleness of pipelines and the safety concerns of pure learning methods, paving the way for a system that is simultaneously safe, adaptive, and efficient.

To provide a clear and comparative overview, the performance and characteristics of the primary motion planning methodologies—including pipeline methods, IL, RL, generative AI, and the proposed data-driven optimal control—are summarized in **Table 4**. This table evaluates each approach against the critical challenges discussed above, offering a high-level snapshot of their respective strengths and weaknesses. The subsequent sections will delve into a detailed analysis of each methodology, elaborating on the insights presented in this summary.

**Table 4** A comparative analysis of motion planning methodologies against key challenges

| Challenge | Pipeline method | IL | RL | Generative AI | DDOC |
|---|---|---|---|---|---|
| Generalization & Adaptability | Challenging | Low | Moderate | High | High |
| | Brittle by design; fails in any scenario not explicitly programmed. | Suffers from covariate shift; fails outside the training data distribution | Limited by the sim-to-real gap and unsafe exploration. | Leverages world knowledge and LLMs' reasoning ability for semantic reasoning on novel scenarios, but risks hallucination. | Using a data-driven approach to adjust internal models and parameters within a structured framework to achieve robust generalization ability. |
| Safety & Explainability | High | Challenging | Challenging | Moderate | High |
| | Fully explainable and verifiable, but safety depends on the completeness of the rules. | "Black box" nature prevents formal verification and makes debugging intractable. | "Black box" nature combined with inherently unsafe trial-and-error learning. | Offers semantic explanations for trust, but lacks formal verifiability and has hallucination risks. | Provides verifiable safety through explicit constraints and maintains high explainability via its control structure. |
| Real-World deployment | High | Moderate | Moderate | Low | Moderate |
| | Computationally | Suffers from a sim- | Sample inefficiency | Computationally | Offer reduced |
| | efficient and highly deployable in real-world vehicles. | to-real gap and requires massive. The training paradigm is relatively mature | and the sim-to-real gap make real-world training and deployment difficult. | prohibitive for real-time control; best suited for high-level guidance. | engineering effort via adaptation, but high computational cost poses challenges for onboard real-time execution. |
| Interactive driving | Moderate | Moderate | High (Theory) / Moderate (Practice): | High | Moderate |
| | Existing rule-based methods can achieve simple interactive driving through prediction. | Can mimic human interactive behaviors but cannot reason about complex scenarios. | Theoretically ideal but practically difficult to train stable multi-agent policies. | Can reason about social context and intent, enabling more sophisticated and human-like interactions. | Provides a mechanism for safe interaction (constraints) but requires a strong external reasoning model. |
| Evolution | Low | Low | High | High | High |
| | The system is static and can only be updated manually via OTA software releases. | Standard offline training results in a fixed policy; online versions are impractical for deployment. | Conceptually built for continuous online learning, but ensuring safety and stability remains an unsolved problem. | Evolution can be supported by integrating offline knowledge and online data. | Its online self-tuning and model adaptation capabilities are a direct mechanism for safe, continuous evolution. |
| Customization | Low | Moderate | Moderate | High | High |
| | Limited to a few pre-defined, hand-tuned parameter sets (e.g., “Sport mode”). | Can learn personalized driving styles by training on data from specific users. But it cannot achieve a finer particle size. | Can be tuned via reward shaping, but mapping preferences to rewards is non-trivial. | Offers intuitive customization through natural language commands and reasoning. | Learning and optimizing user preferences based on a data-driven online adjustment cost function and constraint adaptation. |

## 3. Learning-Based Motion Planning

Pipeline approaches to motion planning increasingly struggle to meet evolving performance requirements. Concurrently, AI technologies have undergone rapid development. Compared to the pipeline methods, AI technologies are highlighted for their capability in adapting to various and complex driving scenarios via learning from empirical data. Consequently, AI presents a promising solution to address the emerging challenges in motion planning. This section discusses three categories of AI-based motion planning methods.

### 3.1. Imitation Learning

IL generates driving policies by imitating expert trajectories. In the IL framework, naturalistic driving data are collected for training, containing state and action pairs. The relation between state and action is modeled as driving policies. An IL problem is formulated as shown in Eq. (2). The objective is to learn a policy $\pi_\theta(s)$ to minimize the approximation error. Then, in the application, policies could be generated according to real-time state $s$.

$$\theta^* = arg\max_{\theta} \sum_{(s^{(i)},a^{(i)})\in\mathcal{D}} log\pi_\theta\left(a^{(i)}\middle|s^{(i)}\right) \tag{2}$$

where $\theta$ represents the training parameters of policy $\pi_\theta(s)$; the state-action pairs $\left(s^{(i)}, a^{(i)}\right)$ denotes the i-th sample in the dataset $\mathcal{D}$; $\pi_\theta\left(a^{(i)}\middle|s^{(i)}\right)$ is the policy parameterized by $\theta$.

IL is generally categorized into three types (Le Mero et al., 2022): Behavioral Cloning (BC), Direct Policy Learning (DPL), and Inverse Reinforcement Learning (IRL). The logic and application of the three approaches are briefly reviewed in this section. **Table 5** presents all prominent IL methods reviewed in this part.

**Table 5** A summary of representative IL-based motion planning methods for AD

| Article | Method type | Input | Output | Scenario | Simulator/Real-world | Problems addressed | Limitations |
|---|---|---|---|---|---|---|---|
| Bojarski (2016) | BC | Monocular RGB | Steering angle | Lane keeping | Real-world | Lateral control via BC | No longitudinal control and high-level navigation; |
| | | | | | | | Lack of interpretability |
| Codevilla et al. (2018) | BC | Monocular RGB; High-level Command+ Measurements | Control actions | Urban driving | Simulator (CARLA) & Physical robot | Ambiguity at intersections | Dataset bias; Generalization; Reactive only |
| Kishky et al. (2024) | BC | Monocular image; High-level command | Steering angle | Urban driving | Simulator (CARLA) | Poor generalization; Regression ambiguity | Complexity; Evaluation scope |
| Lai and Bräunl (2023) | BC | RGB Image | Linear & Angular velocity | Indoor navigation | Real-world | Partially observable environments; Lack of temporal context | Memory drift; Limited to low-speed indoor settings |
| Chen et al. (2021b) | BC | Monocular image | Control actions | Urban driving | Simulator (CARLA) | Interpretability; Sample efficiency | Reconstruction artifacts; Computationally heavy |
| Chen et al. (2020a) | BC | Monocular image; High-level command | Waypoints | Urban driving | Simulator (CARLA) | Sample inefficiency | Teacher dependency; Deployment gap |
| Fan et al. (2024) | BC | Vectorized state representation | Future trajectory | Dynamic traffic environments | Simulator (NuPlan / CARLA) | Safety/Risk; Infeasible trajectories | Conservativeness; Computational cost |
| Chitta et al. (2022) | BC | Monocular Image; LiDAR | Waypoints | Urban driving | Simulator (CARLA) | Ineffective sensor fusion | Computational cost; High latency |
| Hu et al. (2023c) | BC | Multi-camera images | Planning trajectory | Complex urban scenes | Dataset (nuScenes) | Error accumulation; Task isolation | Training resource intensive; Complexity |
| Teng et al. (2022) | BC | Monocular image / Sensor fusion | High-level sub-goals +Low-level Control | Long-horizon navigation tasks | Simulator (CARLA / Similar) | Long-horizon dependency | Sub-optimal performance; Error propagation |
| Cultrera et al. (2020) | BC | Monocular image | Control action + Attention map | Urban & Highway | Dataset (Mapillary / CARLA) | Black-box opacity | Correlation vs Causation; Performance trade-off |
| Sadat et al. (2020) | BC | LiDAR; HD map | Trajectory | Urban driving | Real-world dataset (ATG4D) | Interpretability; Safety | Dependency on HD maps; Complexity |
| Seo et al. (2024) | BC | State observations | Control action | Continuous control | Simulation (MuJoCo / Simple Driving) | Covariate shift | Optimization difficulty |
| Li et al. (2022) | BC | Monocular image (Student) vs. Privileged info (Teacher) | Control actions | Route following (Urban/Country) | Simulator (CARLA) | Sensor limitations; Covariate shift | Teacher quality; Distillation loss |
| Machado and Antonelo (2025) | BC | Observation history | Distribution of future trajectories | Complex interaction | Offline datasets | Multi-modality; Covariate shift | Inference Speed; Real-time constraint |
| Wen et al. (2020) | BC | Sequence of observations | Control actions | Simulation environments (CARLA) | Simulator | Causal confusion | Training instability; Performance drop on simple roads |

| Wen et al. (2021) | BC | Monocular image | Control action | Urban driving | Simulator (CARLA) | Causal confusion; Dataset imbalance | Defining keyframes; Loss of continuity |
|---|---|---|---|---|---|---|---|
| Chuang et al. (2022) | BC | Image sequence + Past action | Residual action | Dynamic driving scenarios | Simulator (CARLA) | Causal confusion | Action drift |
| Ross et al. (2011) | DPL | Monocular image | Control information | Autonomous racing | Simulator (3D racing & navigation environments) | Covariate shift | A constant “on-call” expert |
| Zhang et al. (2021b) | DPL | Monocular image | Control actions | Complex urban driving | Simulator (CARLA) | Sub-optimal experts; | Sim-to-real gap |
| Hoque et al. (2021) | DPL | RGB images | Control actions | Manipulation & Navigation | Simulator & Real-world | High cost of human supervision | Defining thresholds |
| Spencer et al. (2020) | DPL | RGB images & Inertial states | Control commands | Off-road path following | Real-world | Expert burden; Feedback sparsity | Expert vigilance assumption; Recovery bias |
| Perez-Dattari et al. (2022) | DPL | Monocular RGB + human guidance | Trajectory | Unstructured environments | Simulator (PyBullet / Custom Kinematic Sim) | Safety constraints; Interpretability; Covariate shift | Computational latency |
| Lee et al. (2022) | IRL | Occupancy grids | Spatiotemporal costmap | Highway merging & Cruising | Dataset (NGSIM) & Simulator (CARLA) | Black-box policy | Computationally expensive |
| Zhang et al. (2024b) | IRL | Vehicle states & Environment | Trajectory | Urban intersections | Simulation / Dataset | Demonstration ambiguity | Reward ambiguity; Optimization complexity |
| Wu et al. (2020) | IRL | Expert trajectories | Cost function | Highway driving | Simulator (Custom / NGSIM based) | Computational intractability | Local optima; Sample efficiency |
| Wang et al. (2022a) | IRL | Joint state | Diverse policies | Human-Robot interaction | Real-world & Simulator | Mode collapse in GAIL | Training instability |
| Bronstein et al. (2022) | IRL | Vectorized map & Agent history | Future trajectory | Complex urban driving | Dataset (Waymo Open Motion Dataset) | Long-horizon inconsistency; Causal confusion | Model mismatch; Inference latency |
| Jamgochian et al. (2022) | IRL | Vectorized state | Hierarchical controls | Urban driving | Simulator (CARLA / SMARTS) | Safety; Long-horizon planning | Complexity; Conservative behavior |
| Xue et al. (2025) | IRL | Vehicle state | Control Action + Decision intent | driving tasks | Simulator | Training instability of GAIL; understanding of driver intent | Computational cost; Tuning sensitivity |
| Hu et al. (2021) | IRL | LiDAR / Radar / Images | Freespace probability map | urban environments | Datasets (Waymo / Argoverse) | Unknown obstacles | Conservative; Short-horizon |
| Hu et al. (2022) | IRL | Multi-view images | Trajectory | Complex urban scenarios | Dataset (nuScenes) | Temporal inconsistency | High compute; Latency |
| Nan et al. (2025) | IRL | Sequence of vehicle states | Trajectory | Complex interactive scenarios | Simulation / Dataset | Balancing safety with human-likeness | Computational cost; Data requirement |
| Zhang et al. (2023c) | IRL | Surrounding object | Integrated Behavior & Trajectory | Unsignalized Intersections / Merging | Simulator (PreScan / Custom) | Freezing robot problem | Safety risk |
| Hu et al. (2025a) | IRL | User driving data + Environment | Personalized Lane change policy | Highway / Urban lane changing | Simulation / Hardware-in-the-loop | Slow adaptation | Data sparsity; Overfitting |
| | | state | | | | | |
| Hu et al. (2020) | IRL | Monocular image | Steering Angle & Speed | General driving tasks | Dataset (Udacity / Comma.ai) | Task coupling | Error propagation; Latency |
| Teng et al. (2022) | IRL | Monocular image + Vehicle speed | Steering Angle & Speed | Urban driving | Simulator (CARLA) | Interpretability; Sample inefficiency | Mode confusion |
| Zhu and Zhao (2023) | IRL | Ego-vehicle state + Pedestrian states | Longitudinal acceleration | Unsignalized intersections | Simulation (SUMO-based) | Human-vehicle interaction | Perception assumption |

### *3.1.1. Behavioral Cloning*

BC is fundamentally a Supervised Learning (SL) approach. BC based motion planning is generally an end-to-end approach. It directly trains a mapping relation from input perception data to output control commands. This method relies on the collection of abundant expert driving data, which has been regarded as the main challenge. However, BC has been widely adopted in commercial operations, as OEMs and Tier 1 suppliers possess the capacity to acquire such large-scale training datasets. Furthermore, the training process indicates that BC assumes there is an explainable and explicit mapping relation for the expert's actions regarding the environment. However, in reality, a driver's behavior may be influenced by numerous environmental factors, personal preferences, and even stochasticity, which are all latent variables that cannot be comprehensively collected and modeled.

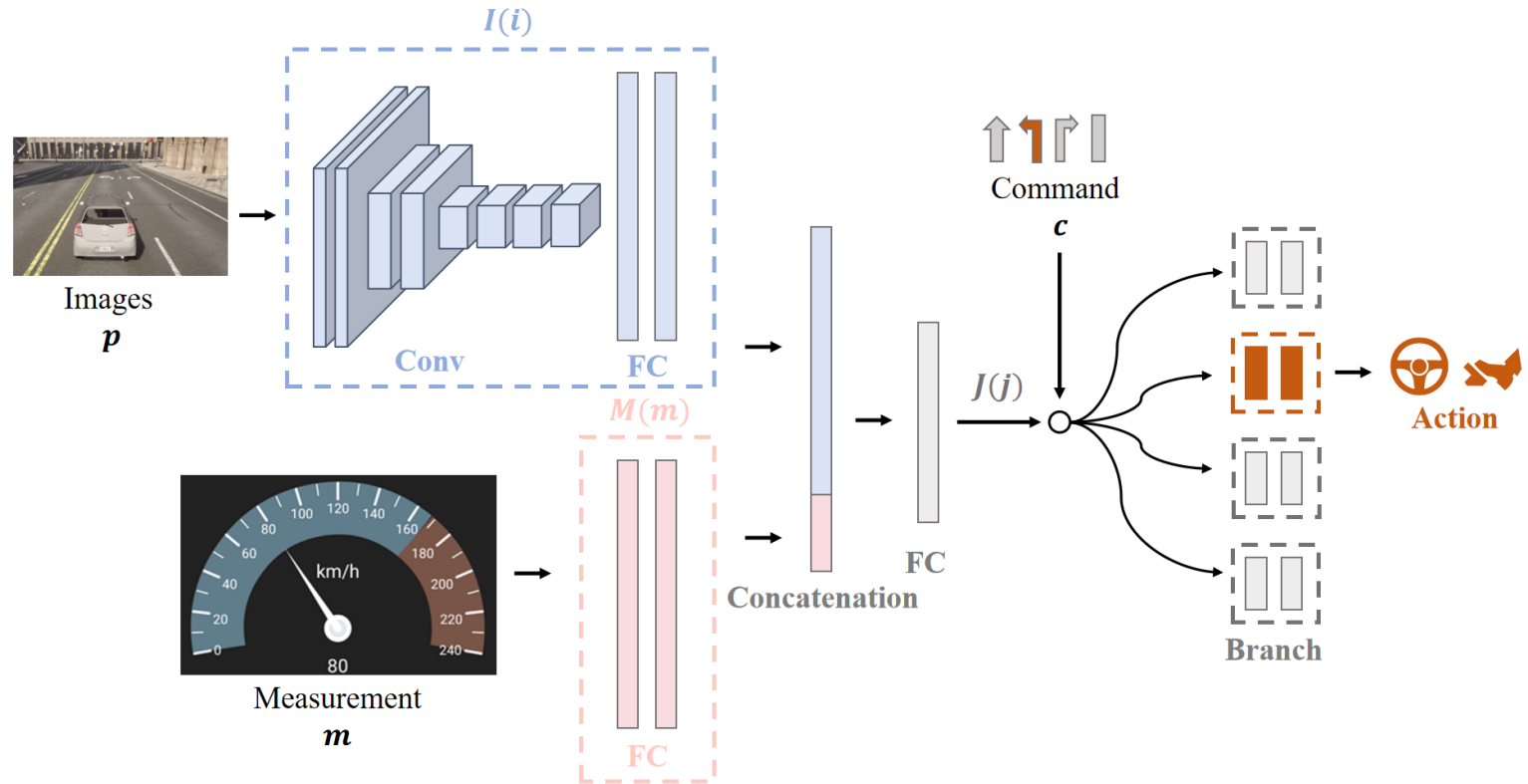

**Fig. 4.** CIL architecture from (Codevilla et al., 2018)

From a methodological perspective, Bojarski et al. (Bojarski, 2016) pioneered the concept of end-to-end AD by utilizing Convolutional Neural Networks (CNNs) to map raw image inputs directly to control commands, thereby validating the potential of learning-based approaches for implicit feature extraction. However, purely perception-based mapping suffers from "navigational ambiguity" in complex scenarios such as intersections. To address this limitation, Codevilla et al. (Codevilla et al., 2018) introduced Conditional Imitation Learning (CIL) (as shown in Fig. 4). By incorporating high-level navigational commands as conditional inputs, this approach effectively decoupled high-level decision-making from low-level control, establishing a foundational architecture for subsequent BC methodologies. Nevertheless, standard BC often lacks robustness in complex dynamic environments. To tackle this, Chen et al. (Chen et al., 2020a) proposed the "Learning by Cheating" (LBC) paradigm (Fig. 5). This method diverges from traditional imitation of human data by adopting a distillation strategy where a "privileged agent" guides a "sensorimotor agent." By enabling the student model to mimic an expert with omniscient views, LBC significantly improved closed-loop driving performance. At the architectural level, as operational domains expand to complex urban intersections, single-modality perception struggles to meet safety and performance requirements, while traditional CNNs often fail to effectively fuse global features from multimodal data. TransFuser (Chitta et al., 2022) (Fig. 6) introduced a significant improvement by leveraging the self-attention mechanism of Transformers to achieve deep fusion of LiDAR point clouds and camera imagery across multiple feature scales. This architectural innovation effectively resolved complex interaction challenges, offering a novel paradigm for high-dimensional sensor fusion in end-to-end models.

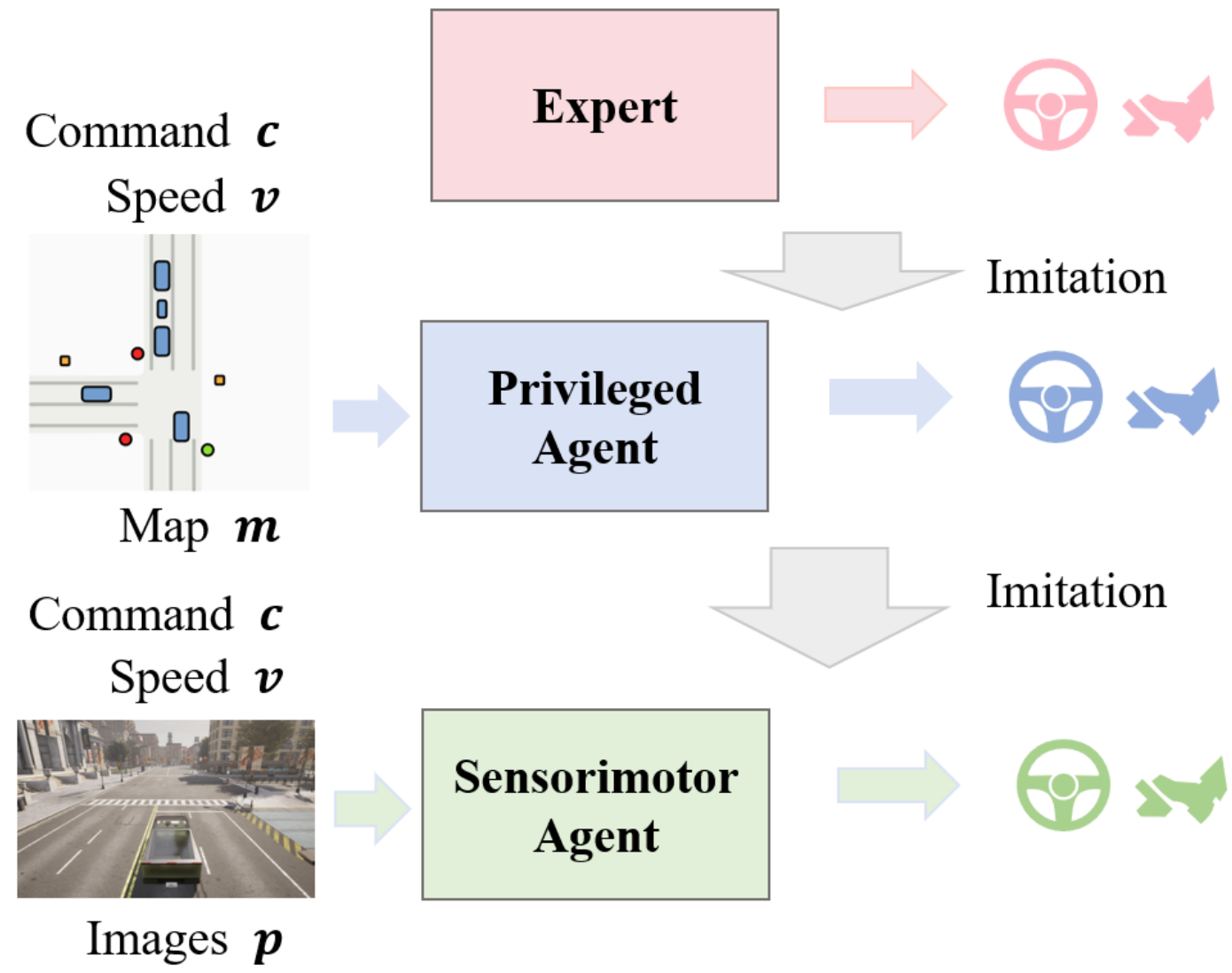

**Fig. 5.** Overview of LBC model (Chen et al., 2020a). Privileged agents can obtain privileged information (Bird's-Eye View (BEV)) like cheating, and then learn expert robust strategies; Sensorimotor Agent is an agent that cannot obtain privileges. By learning to mimic Privileged Agent (similar to a white box that can provide online policy supervision), it ultimately obtains an intelligent agent that does not cheat.

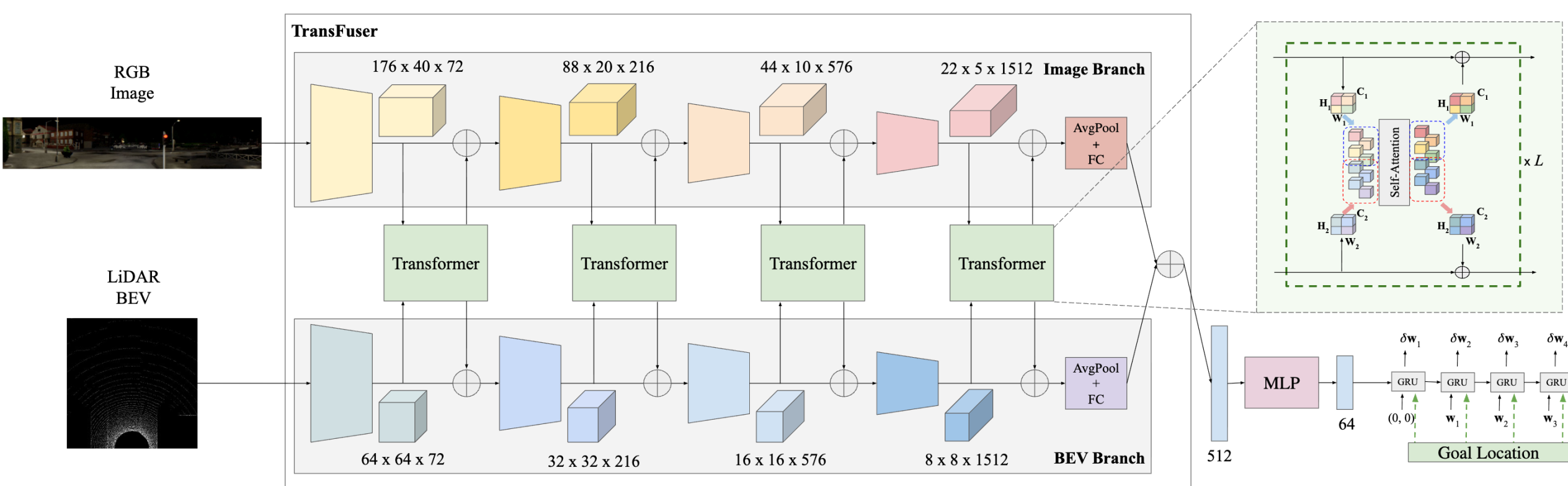

**Fig. 6.** TransFuser architecture of the transfuser model (Chitta et al., 2022). The model contains two parallel processing branches. Within the encoders of these branches, feature maps at different hierarchical levels (i.e., features at varying resolutions) interact and are fused through Transformer modules. The fused features are then passed into an autoregressive network to generate future trajectory predictions.

Despite these significant advancements, the real-world deployment of BC remains constrained by three fundamental challenges. The first is the lack of interpretability stemming from the "black-box" nature of end-to-end models. Hu et al. (Hu et al., 2023c) provided a pioneering direction with UniAD (**Fig. 7**), the industry's first planning-oriented framework unifying full-stack tasks. By outputting explicit intermediate representations (e.g., object detection, trajectory prediction), UniAD demonstrated that modular end-to-end learning can achieve both high performance and interpretability. Additionally, other studies have sought to address this issue by building hierarchical architectures (Teng et al., 2022), utilizing attention models (Cultrera et al., 2020), or introducing intermediate representations (Sadat et al., 2020), thereby improving model interpretability and driving safety to a certain extent. The second challenge is covariate shift (Seo et al., 2024), arising from the mismatch between the state distributions induced by the learned policy and the expert policy. To mitigate this, recent works such as Diffusion-BC (Machado and Antonelo, 2025) employed diffusion models to capture multi-modal behavior distributions, enhancing generalization in unseen scenarios. Similarly, Li et al. (Li et al., 2022) utilized task knowledge distillation to transfer driving policies across scenarios, further improving scenario generalization. The final challenge is causal confusion (Muller et al., 2005), where the agent learns spurious correlations between inputs and outputs. Researchers have explored various mitigation strategies, such as employing adversarial training to eliminate spurious temporal correlations (Wen et al., 2020), increasing the weight of keyframes in the loss function (Wen et al., 2021), and supervising networks using action residuals instead of raw actions (Chuang et al., 2022). While these methods have yielded improvements in specific scenarios, causal confusion remains a significant hurdle for BC.

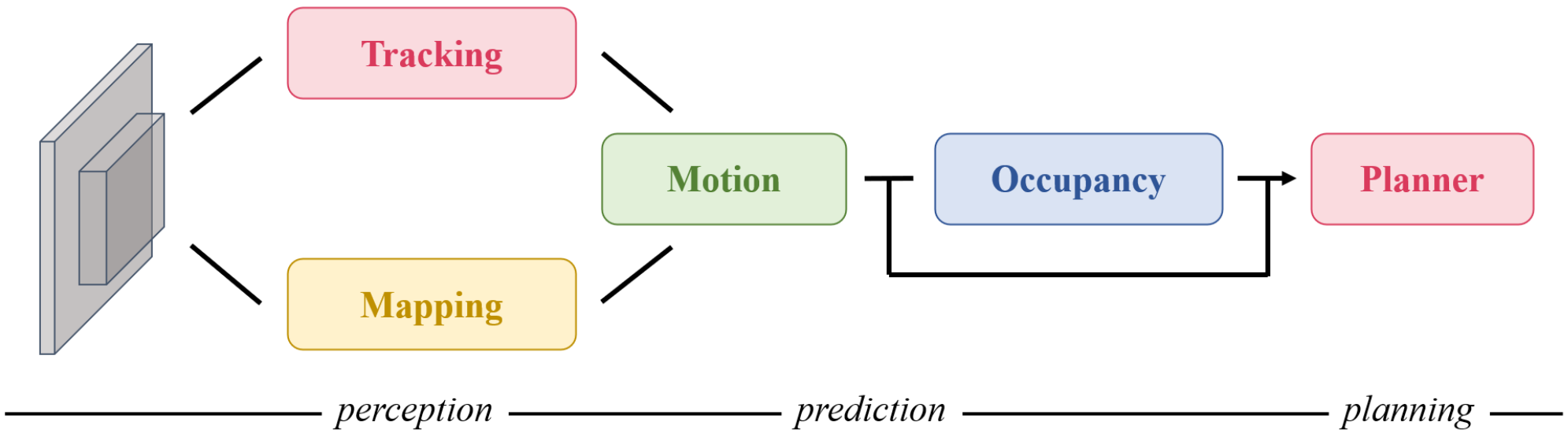

**Fig. 7.** Overview of the UniAD (Hu et al., 2023c). For the first time, the three main tasks of perception, prediction, and planning, as well as six sub-tasks of object detection, object tracking, scene mapping, trajectory prediction, grid prediction, and path planning, have been integrated into a unified end-to-end network framework, achieving a universal driving model for key tasks across the entire stack

*3.1.2. Direct Policy Learning*

DPL is an interactive method that learns the desired optimal behavior by directly querying experts (Luo et al., 2015). This approach addresses the issue of insufficient state information in expert datasets, as it can access experts in real-time. Compared to BC, the primary advantage of DPL lies in leveraging expert trajectories to guide the agent in recovering from trajectory deviations (Attia and Dayan, 2018). However, while real-time expert querying is feasible in simulation, it incurs high costs in real-world applications (Sumanasena et al., 2023), thereby limiting the method's scalability in practical scenarios.

From a methodological perspective, Ross et al. (Ross et al., 2011) introduced a seminal IL method called Dataset Aggregation (DAgger). In IL, the learner's predictions influence future observations, potentially placing the learner in previously unseen states and compounding errors. DAgger mitigates these issues by iteratively collecting data and updating the policy. However, recent advancements (2020–2025) have moved beyond the original DAgger pipeline—which suffers from query inefficiency and safety concerns—toward two advanced paradigms: *privileged automated supervision* and *intervention-based learning*. On one branch, recognizing that humans are "good drivers but poor coaches" for dense on-policy supervision (due to latency and inconsistency), a prevailing trend is to replace human demonstrators with automated oracles. For instance, Roach (Zhang et al., 2021b) trains a RL coach with privileged BEV inputs. The sensor-based student policy then distills this robust expert via DAgger. This "privileged distillation" strategy effectively decouples supervision quality from human limitations, enabling the imitation agent to achieve super-human generalization on challenging benchmarks like CARLA. On the other branch, parallel to automated experts, the mechanism of human interaction has evolved from "active querying" (where the robot solicits labels) to Learning from Interventions. In this paradigm, the human expert functions as a safety guard, taking over control only when the agent fails. ThriftyDAgger (Hoque et al., 2021) formalizes this by introducing budget-aware gating based on novelty and risk, ensuring the agent only switches to the expert in unfamiliar or critical states. Similarly, Expert Intervention Learning (Spencer et al., 2020) treats the absence of human intervention as an implicit signal that the current policy is "good enough," thereby focusing the costly supervision solely on failure recovery. This shift significantly reduces the annotation burden while specifically targeting the most critical long-tail scenarios. Finally, to ensure safety during the interactive exploration phase, recent works integrate control theory into DPL. For example, Social-MPCC (Perez-Dattari et al., 2022) combines interactive imitation with Model Predictive Contouring Control, using the controller to enforce kinematic and safety constraints while aggregating demonstrations.

In summary, modern DPL methods have moved beyond simple imitation, evolving into structured frameworks that leverage privileged automated coaches and sparse human interventions to address the efficiency and safety bottlenecks of on-policy learning.

*3.1.3. Inverse Reinforcement Learning*

IRL addresses the limitations of BC, which mimics expert behavior without reasoning about the underlying motivations. While both methods begin by collecting expert trajectories, IRL diverges from BC by inferring the underlying reward function structure rather than mapping states directly to actions. The behavioral policy is then optimized based on this inferred reward. This approach circumvents the difficulty of manually defining reward functions for complex AD tasks—a significant challenge in traditional RL. Nevertheless, IRL faces several hurdles, including training instability and high computational costs. Furthermore, the interpretability of the model and the high demand for data are also important challenges faced by IRL.

As early as 2004, Abbeel and Ng (Abbeel and Ng, 2004) proposed feature-based IRL, whose basic idea is to find a reward function so that all agents generate rewards smaller than the expert's reward, and then use this reward function to train agents. The core of the algorithm is to use the maximum margin method to calculate the parameter value of the current reward function in the optimal strategy obtained by iteration. This part can be expressed as Eq. (3):

$$\begin{aligned} &\max_{t,w} m \\ &\text{s.t. } w^{\mathrm{T}}\mu_E \geqslant w^{\mathrm{T}}\mu^{(j)} + t, j = 0, \cdots, i-1 \\ &\quad \| w \|_2 \leq 1 \end{aligned} \tag{3}$$

Where $\mu^{(j)}$ denotes the feature expectation of the optimal policy obtained in the j-th iteration, $\mu_E$ represents the characteristic expectation of the expert strategy, $m$ is the margin of the two strategies. $w$ is the weight of feature expectation in the reward function. This method used a feature-based IRL method to learn different driving styles in highway simulation, and achieved promising results. However, the main issue with this method is the ambiguity of the reward function (Ng and Russell, 2000). To mitigate this, Ziebart et al. (Ziebart et al., 2008) extended the feature-based IRL method and proposed the Maximum Entropy IRL approach. This method effectively reduces ambiguity caused by multiple possible reward weights. Subsequently, many studies (Lee et al., 2022; Wu et al., 2020; Zhang et al., 2024b) applied the maximum entropy method to end-to-end AD tasks. Building on the maximum entropy method, Ho et al. (Ho and Ermon, 2016) proposed the Generative Adversarial Imitation Learning (GAIL) algorithm. The core idea of this algorithm is to directly extract strategies from data through generative adversarial methods, thereby overcoming the indirectness and inefficiency issues of traditional IRL methods. The GAIL algorithm has become a classic method in this field, inspiring numerous variants such as Co-GAIL (Wang et al., 2022a), MGAIL (Bronstein et al., 2022), SHAIL (Jamgochian et al., 2022), and AWGAIL (Xue et al., 2025), which have achieved significant results in their respective application areas.

Despite these advancements, IRL in end-to-end AD faces persistent challenges regarding safety, interpretability, and generalization. To enhance transparency, recent studies (Hu et al., 2021; Hu et al., 2022; Nan et al., 2025) have integrated auxiliary perception tasks into cost function optimization, significantly improving decision-making safety. In interactive scenarios, Zhang et al. (Zhang et al., 2023c) utilized online IRL to infer personalized cost functions within a unified behavior-motion planner, enhancing efficiency in complex maneuvers. Addressing the data dependency of personalization, Hu et al. (Hu et al., 2025a) proposed a "Lesson Learning" framework that leverages apprenticeship learning during user takeovers to iteratively refine safety constraints. Furthermore, to mitigate the inherent limitations of expert datasets (Hu et al., 2020) and the risk of "unknown state" failures, researchers (Teng et al., 2022; Zhu and Zhao, 2023) have employed data augmentation and sim-to-real synthesis to enrich data distribution and improve model robustness.

### 3.2. Reinforcement Learning

RL is an important machine learning paradigm in which an agent learns an optimal strategy through iterative interactions with an environment. RL is usually modeled by Markov Decision Processes (MDPs), which are composed of state space $S$, action space $A$, reward $R: S \times A \rightarrow R$, state transition function $T: S \times A \times S \rightarrow [0, 1]$ and time discount factor $\gamma$. The agent takes action $a \epsilon A$ according to the policy $\pi: S \times A \rightarrow [0, 1]$ in different states $s \epsilon S$ of the environment, and the environment gives the corresponding reward $r \epsilon R$ according to the agent's action $a$ and transfers to the new state according to the transfer function $T$. The agent's objective is to maximize the cumulative reward, as formulated in Eq. (4), where $\pi^*$ is the optimal strategy, $H$ is the number of time steps, reward $r_{t+1} \epsilon R$, initial state $s_0 \epsilon S$ and the discount factor $\gamma \epsilon [0, 1]$. The agent gradually builds a deep understanding of the environment through trial and error, so as to master the ability to make the best choice in a variety of complex situations.

$$\pi^* = \underset{\pi}{arg\,max}\, \mathbb{E}_\pi \left[ \sum_{t=0}^{H-1} \gamma^t r_{t+1} \mid s_0 = s \right] \tag{4}$$

Owing to its inherent learning characteristics, RL-trained agents demonstrate superior generalization in novel scenarios compared to Imitation Learning (IL), which often struggles with distribution shifts (Chib and Singh, 2023). However, RL frequently faces challenges regarding insufficient exposure to diverse driving scenarios during training—a limitation less prevalent in IL. In addition, the data efficiency of RL is lower than that of IL, and its use in the real world is also challenging (Tampuu et al., 2020).

To systematically review these approaches, we categorize existing methodologies based on action spaces and learning paradigms. We first introduce discrete and continuous action methods, followed by a discussion on hybrid learning methods that integrate RL with other paradigms to address inherent limitations in sample efficiency and stability. **Table 6** presents all prominent RL methods reviewed in this part.

**Table 6** A summary of representative RL-based motion planning methods for AD

| Article | Method type | Input | Output | Scenario | Simulator/Real-world | Problems addressed | Limitations |
|---|---|---|---|---|---|---|---|
| Wolf et al. (2017) | DQN | Occupancy grid map | Discrete steering actions | Lane keeping & Obstacle avoidance | Simulator (VTD - Virtual test Drive) | Feature abstraction | Jerky Control; Lack of longitudinal control; Sim-only |
| Chen et al. (2020b) | DQN | State vector; High-level command | Discrete motion commands | Highway overtaking & Lane changing | Simulator (SUMO) | Reward shaping | Discrete actions; Rule dependency |
| Zhao et al. (2024a) | DQN | Reference trajectory points; Vehicle state | Discrete control | Path tracking | Simulator (CarSim + Simulink) | Tracking accuracy | Oscillation risk |
| Jin et al. (2025) | DQN | Vehicle state; Surroundings | Discrete behavior + Continuous parameters | Dynamic traffic | Simulator (SUMO / Python-based) | Action space limitation; Conflicting objectives | Reward tuning; Computational complexity |
| Li et al. (2024b) | DQN | Interaction features | Discrete action + Continuous args | Interactive highway driving | Simulator (SUMO / CARLA) | Conservative behavior | Prediction uncertainty; Training stability |
| Ehrhardt et al. (2025) | GCM-DQN | State observation | Optimized continuous parameters | Planning tasks | Benchmark environments / Simulator | Sub-optimal parameters | Inference latency; Local optima |
| Osiński et al. (2020) | PPO | Monocular RGB image | Continuous control | Lane following / Track driving | Simulator (CARLA); Real-world | Sim-to-real gap | Scenario simplicity; Visual dependence |
| Guan et al. (2020) | PPO | Joint state of all vehicles | Joint accelerations for all vehicles | Non-signalized intersection | Simulator (Numerical / SUMO-based) | Traffic efficiency | Communication delay; Scalability |
| Wu et al. | PPO | Sensor states | Continuous | Highway / | Simulator | Value | Complexity |

| (2021a) | | | control | Obstacle avoidance | (TORCS) | overestimation | |
|---|---|---|---|---|---|---|---|
| Ye et al. (2020) | PPO | Surrounding vehicles info | Lane change decision | Highway lane changing | Simulator (VISSIM / SUMO) | Policy stability | Interaction limitations; Safety guarantee |
| Zhang et al. (2022) | PPO | Vehicle state + Local map | Sequence of control actions | Parking / Obstacle avoidance | Simulation (MATLAB / Python) | Myopic behavior | Computation; Horizon tuning |
| Jin et al. (2024a) | HRL | State observation | High-level goal + Low-level trajectory | Complex maneuvers | Simulator (SUMO / CARLA) | Unstable control; Sparse rewards | Complexity; Constraint rigidity |
| Delavari et al. (2025) | PPO | State observation | Masked action set | General driving tasks | Simulation benchmarks | Curse of dimensionality; Slow convergence | Bias; Generalization |
| Hu et al. (2024b) | HCPI-RL | Continuous driving data | Policy update | Continuous learning | Simulator | Performance degradation; Catastrophic forgetting | Conservative updates |
| Zhou et al. (2023) | A3C | Vehicle states; Surroundings | Discrete lane change decision | Highway lane change | Simulator (SUMO) | Training efficiency | Discrete decisions; Data efficiency |
| Duan et al. (2024) | SAC | Vectorized state | Continuous control | Multi-lane highway driving | Simulator (SUMO) | Risk assessment; Exploration | Computational complexity |
| Kendall et al. (2019) | DDPG | Monocular image | Continuous control | Lane following | Real-world | Sim-to-real gap; Data efficiency | Safety risk; Simple task |
| Saxena et al. (2020) | TD3 | Occupancy grid | High-level actions | Dense traffic | Simulator (SUMO / Custom) | Crowded environments | Reward shaping |
| Liang et al. (2018) | IL+DDPG | Monocular image + High-level command | Continuous control | Urban driving | Simulator (CARLA) | Sample inefficiency; Dead-end problem | Reward engineering; Mode confusion |
| Wu et al. (2022) | IL+TD3 | Vehicle states + Human intervention | Continuous control | Highway / Urban lane keeping | Simulator (TORCS / Custom) | Exploration safety; Data efficiency | Human burden; Teacher bias |
| Hu et al. (2023b) | IL+PPO | V2X info | Acceleration / Steering | Ramp merging / Congested traffic | Simulator (SUMO) | Partial observability; Data efficiency | V2X dependency |
| Huang et al. (2022) | IL+SAC | Sensor features | Continuous control | Complex urban driving | Simulator (CARLA) | Unsafe exploration; Unnatural behaviors | Expert quality; Conservative |
| Chen et al. (2021a) | IL+PPO | Monocular image | Waypoints | Urban driving (CARLA Leaderboard) | Learned simulator | Sim-to-real gap; Data interactivity | Compounding errors |
| Zhang et al. (2021b) | IL+PPO | BEV (Privileged) / Image | Control actions | Complex urban driving | Simulator (CARLA) | Expert suboptimality | Teacher upper bound |
| Huang et al. (2025b) | VLM+RL | Image + Text prompts | Continuous control | Risk scenarios / Safety-critical driving | Simulator (Highway-env) | Semantic gap | Inference latency; Hallucination |
| Xu et al. (2025) | LLM+RL | State description + LLM prompts | Control actions | Interactive driving | Simulator (Highway-env / CARLA) | Reasoning deficit in RL; Real-time constraint of LLMs | Synchronization; cost; Alignment |

### *3.2.1. Discrete Action Methods*

Methods designed for discrete action spaces are predominantly founded upon Value-Based RL. In this paradigm, an agent indirectly learns the optimal policy by learning the optimal value function. This approach is often simple, intuitive, and offers stable performance, especially excelling in discrete action spaces. However, these methods face limitations in addressing stochastic policies. Furthermore, given that end-to-end AD systems typically necessitate precise continuous control, the applicability of discrete methods in this domain remains restricted (Sharma and Nagpal, 2024).

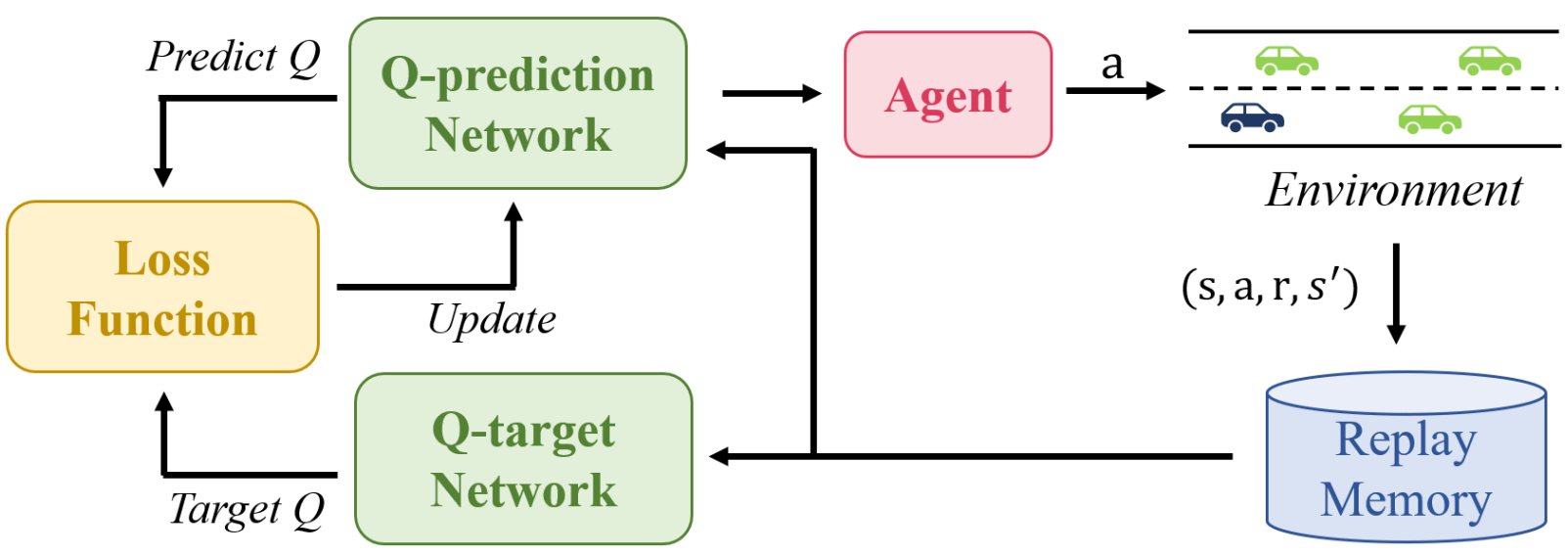

**Fig. 8.** Architecture of DQN-based end-to-end AD method

The most representative Value-Based Method is Q-learning (Watkins and Dayan, 1992). It is a model-free Temporal Difference (TD) algorithm that learns the utility of each state-action pair (the Q-function in Eq. (5)). Deep Q Networks (DQN) (Fig. 8) enhance the model's applicability by using Deep Neural Networks (DNNs) as non-linear Q-function approximators for high-dimensional state spaces.

$$Q(s,a) \leftarrow Q(s,a) + \alpha[r + \gamma \max Q(s',a') - Q(s,a)] \quad (5)$$

Here, $Q(s,a)$ represents the value estimation of action $a$ selected in state $s$, $s'$ denotes the next state after executing action $a$ in state $s$; $a'$ represents a candidate action in the next state $s'$. The operator "←" indicates the update of $Q(s,a)$ towards the target value, $\alpha \epsilon [0,\ 1]$ is the learning rate that controls the update degree of Q value in each time step, and $\gamma \epsilon [0,\ 1]$ is the discount factor. Early adaptations, such as Wolf et al. (Wolf et al., 2017), demonstrated the potential of DQN in steering control. To overcome the instability arising from complex perception inputs, recent variations have been proposed. For instance, Chen et al. (Chen et al., 2020b) introduced Conditional DQN to resolve motion command dependencies, while Zhao et al. (Zhao et al., 2024a) leveraged Double DQN to enhance trajectory tracking accuracy beyond conventional controllers.

Despite these structural improvements, the fundamental reliance on fixed discrete actions poses inherent challenges for the naturally continuous domain of AD. A primary limitation lies in the trade-off between control smoothness and computational efficiency. Coarse discretization often leads to oscillatory and jerky trajectories that compromise passenger comfort and vehicle stability (Mysore et al., 2021). Conversely, increasing the resolution to achieve smoother control inevitably triggers the curse of dimensionality, exponentially expanding the action space and severely hampering the convergence speed of Value-Based algorithms (Delavari et al., 2025). To bridge the gap between discrete decision logic and continuous execution, contemporary research has increasingly shifted towards Hybrid Parameterized Action spaces. Unlike traditional methods that treat all actions as independent, discrete options, these approaches—often implemented as Parameterized DQN—decompose the control task into a high-level discrete decision coupled with continuous low-level parameters. Recent frameworks, such as HPA-MoEC (Jin et al., 2025) and HPA-IDRL (Li et al., 2024b), have successfully integrated this hybrid architecture to achieve smooth, interaction-aware behaviors that outperform traditional discrete baselines. Similarly, advancements in Goal-Conditioned Model-Augmented DQN (GCM-DQN) (Ehrhardt et al., 2025) have further enhanced planning capabilities within these parameterized spaces.

*3.2.2. Continuous Control Methods*

Methods designed for continuous action spaces are essential for the precise operational control of autonomous vehicles. These approaches typically employ the Actor-Critic architecture, which combines the advantages of policy optimization and value estimation. Specifically, the actor is responsible for learning the policy, while the critic learns the value function to evaluate the actor's actions. Since the critic can see the potential rewards of a given state, it provides feedback to update the actor's actions step by step, which significantly improves efficiency compared to algorithms that can only update after each episode (e.g., Policy Gradients). However, due to the involvement of two networks and continuous parameter updates in the state space, the correlation between parameters before and after each update leads to the network having a partial view of the problem, thus reducing learning effectiveness. Based on the policy characteristics and update mechanisms, these methods can be broadly categorized into stochastic policy methods and deterministic policy methods.

Stochastic policy methods output a probability distribution over actions (typically a Gaussian distribution), inherently facilitating exploration and enhancing robustness against sensor noise. Among them, Proximal Policy Optimization (PPO) (Schulman et al., 2017) serves as a dominant baseline due to its balance between ease of tuning and performance. Early adopters validated PPO's efficacy in sim-to-real transfer (Osiński et al., 2020) and safe lane-changing tasks by integrating safety intervention modules (Guan et al., 2020; Wu et al., 2021a; Ye et al., 2020). To further address the computational burden and stability issues in complex scenarios, researchers have focused on architectural optimizations. Zhang et al. (Zhang et al., 2022) incorporated PPO into a Receding-Horizon framework to decompose long-term planning into solvable subproblems, while Jin et al. (Jin et al., 2024a) proposed a Hierarchical Reinforcement Learning (HRL) framework that decouples discrete high-level intentions from low-level continuous trajectories to achieve stable long-term planning. Addressing training convergence, Delavari et al. (Delavari et al., 2025) introduced dynamic masking strategies to accelerate convergence in large action spaces, and Hu et al. (Hu et al., 2024b) developed the High-Confidence Policy Improvement Reinforcement Learning (HCPI-RL) framework with a confidence discriminator to ensure monotonic policy improvement. Similarly, for the Asynchronous Advantage Actor-Critic (A3C) algorithm (Mnih et al., 2016), Zhou et al. (Zhou et al., 2023) improved the update mechanism to eliminate convergence oscillations. Furthermore, Soft Actor-Critic (SAC) extends the concept of stochasticity by maximizing policy entropy to encourage exploration. Leveraging this property, Duan et al. (Duan et al., 2024) proposed a distributional SAC framework that effectively captures complex interactions in dense, dynamic traffic, addressing limitations where traditional constraints might fail.

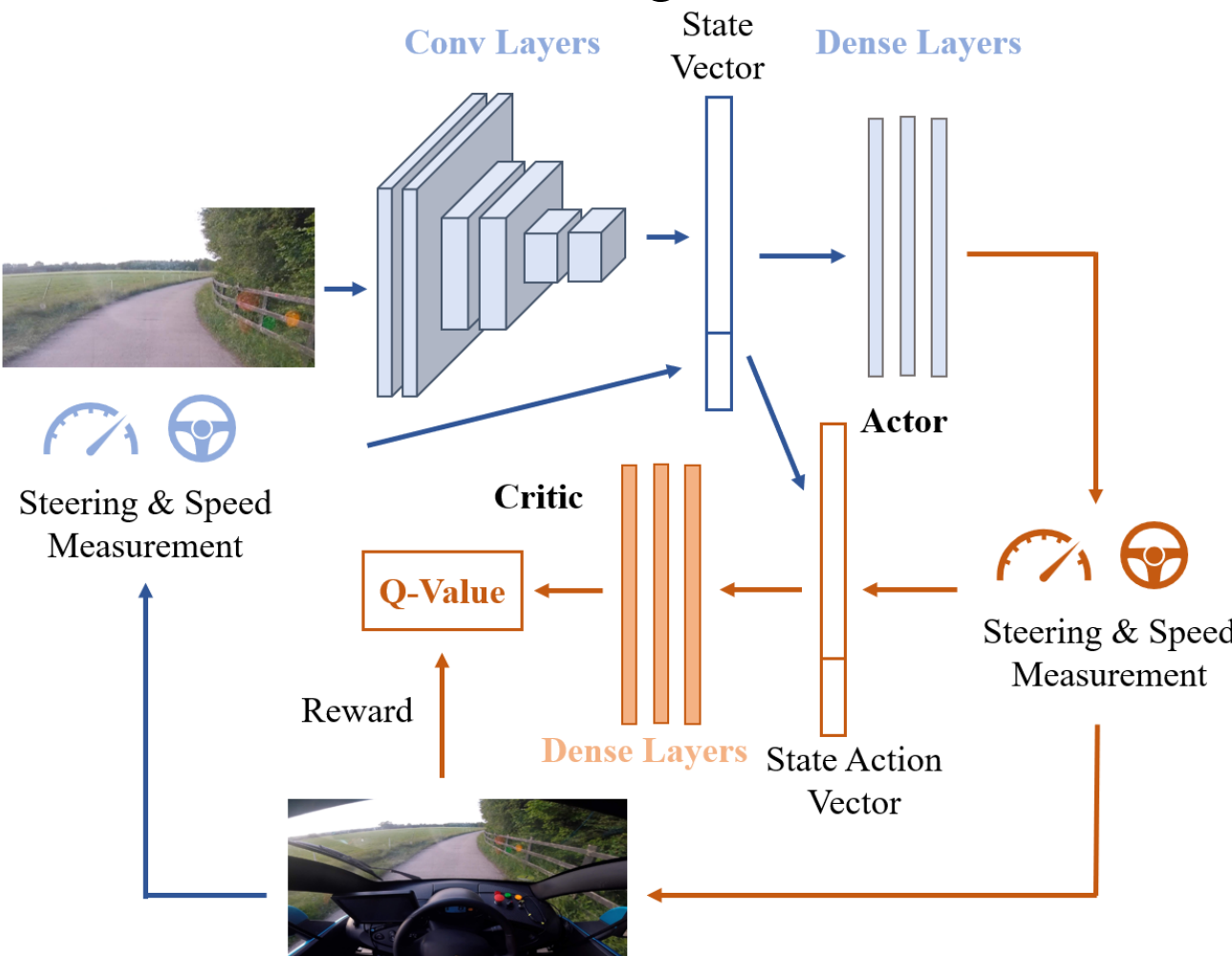

**Fig. 9.** An end-to-end AD framework based on DDPG algorithm (Kendall et al., 2019).

Deterministic policy methods map states directly to specific action values, enabling the use of off-policy learning with replay buffers to significantly improve sample efficiency compared to stochastic methods. Deep Deterministic Policy Gradient (DDPG) (Lillicrap, 2015) is the pioneering algorithm in this category. Kendall et al. (Kendall et al., 2019) marked a milestone in the field by applying DDPG to a full-scale autonomous vehicle firstly (Fig. 9). By adding discrete Ornstein-Uhlenbeck noise to the DDPG algorithm, they significantly improved its robustness, proving the feasibility of Deep Reinforcement Learning (DRL) for physical control tasks. However, DDPG is prone to Q-value overestimation, leading to suboptimal policies. To mitigate this, Saxena et al. (Saxena et al., 2020) adopted the Twin Delayed

DDPG (TD3) algorithm (Fujimoto et al., 2018), which employs clipped double Q-learning to enhance stability. Their experiments in the CARLA simulator demonstrated that TD3 could implicitly learn complex vehicle interactions in congested scenarios, outperforming traditional Model Predictive Control (MPC) in terms of decision-making efficiency and planning robustness.

#### *3.2.3. Hybrid Learning Methods*

While pure RL algorithms offer theoretical optimality, they suffer from severe sample inefficiency and cold-start safety risks in real-world deployment (Horgan et al., 2018). Exploring from scratch in complex traffic scenarios is computationally expensive and potentially dangerous. To mitigate these limitations, hybrid learning methods have emerged, aiming to inject prior knowledge—ranging from human demonstrations to semantic reasoning—into the RL framework.

A dominant strategy for enhancing data efficiency involves bootstrapping the RL agent using expert demonstrations. Liang et al. (Liang et al., 2018) proposed Controllable Imitation Reinforcement Learning (CIRL), a two-stage framework that initializes the policy network via SL based on human driving data, followed by DDPG-based fine-tuning. This "pre-train then fine-tune" paradigm effectively bypasses the tedious initial exploration phase. Building on this, Wu et al. (Wu et al., 2022) introduced a human-guided framework utilizing Prioritized Experience Replay (PER) to prioritize high-value human interventions, while subsequent studies (Hu et al., 2023b; Huang et al., 2022) confirmed that combining IL constraints with RL objectives significantly accelerates convergence compared to pure RL. Conversely, some approaches leverage RL to enhance the robustness of SL agents, addressing the "distribution shift" limitation of IL. Chen et al. (Chen et al., 2021a) utilized RL to estimate action values via the Bellman equation, transferring this value knowledge to a visual motion policy via supervision. Similarly, Zhang et al. (Zhang et al., 2021b) trained an RL expert to generate high-quality supervision signals, enabling the IL agent to learn better generalization capabilities than could be derived from limited human data alone.

Beyond integration with IL/SL, recent advancements have incorporated LLMs and Vision-Language Models (VLMs) to address challenges in reward engineering and high-level reasoning. Designing dense reward functions for complex urban driving is notoriously difficult. Huang et al. (Huang et al., 2025b) introduced the VLM-RL framework, which innovatively employs a pre-trained VLM to interpret driving scenes and compute semantic rewards, replacing brittle manual reward rules. Furthermore, to combine semantic understanding with real-time control, Xu et al. (Xu et al., 2025) proposed a hierarchical "fast-slow" framework. In this architecture, an LLM handles high-level command parsing (the "slow" logical reasoning), while an actor-critic RL agent executes low-level dynamic control (the "fast" reflex), achieving superior performance by leveraging the strengths of both paradigms.

### 3.3. Generative AI

Generative AI refers to machine learning architectures that generate novel data instances by identifying patterns and relationships within training datasets (Feuerriegel et al., 2024). Unlike traditional AI systems that execute specific tasks based on predefined rules, generative models create content—including text, images, video, and audio—that is often indistinguishable from human-generated data. Fundamentally, generative AI emphasizes creativity, enabling machines to simulate the creative process, thus opening new frontiers in innovation. Notable examples include OpenAI's ChatGPT for text generation, DALL-E for image generation, and AIVA for music composition. **Table 7** summarizes the prominent generative AI methods reviewed in this section.

**Table 7** A summary of representative generative AI-based motion planning methods for AD

| Article | Method type | Input | Output | Scenario | Simulator/Real-world | Problems addressed | Limitations |
|---|---|---|---|---|---|---|---|
| Zheng et al. (2024a) | VAE | Multi-view image sequence | Planning trajectory; Future map | Complex urban driving | Dataset (nuScenes) | Uncertainty modeling | Inference latency; Hallucination |
| Salzmann et al. (2020) | CVAE | Agent history, Map data | Probabilistic future rrajectories | Crowded spaces | Dataset (nuScenes / ETH / UCY) | Mode collapse; Dynamic feasibility | Discrete odes; Short horizon |
| Jiang et | Diffusion | Map; Agent | Joint future | Waymo Open | Dataset (Waymo) | Consistency; | Inference speed; |

| al. (2023) | Model | histories | trajectories of all agents | Motion Dataset | | Controllability | Sampling stochasticity |
|---|---|---|---|---|---|---|---|
| Huang et al. (2025c) | Diffusion + PPO | State / Observation | Action / Trajectory | Complex planning tasks | Simulator (nuPlan) | Sub-optimal demonstrations; Distribution shift | Training complexity; Reward alignment |
| Jiang et al. (2024) | Diffusion Model | Partial scene; Text prompts | 3D Scene; Dynamic trajectories | Simulation generation | Dataset (Waymo) | Data scarcity; Scene realism | Logical consistency; Real-time generation |
| Fu et al. (2024) | LLM | Structured environment text | Decision; explanation | Closed-loop simulation | Simulator (CARLA) | Interpretability; Long-tail reasoning | Inference latency; Hallucination |
| Yang et al. (2025a) | LLM | 3D LiDAR point clouds; Text prompt | 3D scene description | 3D scene understanding | Dataset (nuScenes) | Modality gap; Zero-shot 3D recognition | Resolution loss; Spatial hallucination |
| Tang et al. (2025) | LLM | Natural language query | Retrieved BEV scene | Scenario retrieval | Dataset (nuScenes) | Data accessibility; Semantic search | BEV distortion; Query complexity |
| Park et al. (2024) | LLM | Image; Semantic attributes | Pedestrian bounding boxes | Pedestrian detection (driving) | Dataset (CityPersons) | Visual ambiguity; Robustness | Attribute dependency; Real-time cost |
| Elhafsi et al. (2023) | LLM | Description of state & action | Anomaly score | Task planning | Simulator & Real robot | Semantic errors | Description bottleneck; Latency |
| Chib and Singh (2025) | LLM | Pedestrian history; Scene context | Future trajectory | Pedestrian prediction (Urban) | Datasets (ETH / UCY) | Social intent understanding | Inference speed |
| Keysan et al. (2023) | LLM | Trajectory; Scene text description | Future trajectory | Traffic prediction (nuScenes) | Dataset (nuScenes) | Contextual ambiguity | Modality alignment; Generation cost |
| Zhang et al. (2023a) | LLM | Natural language query | Route | Navigation / Trip planning | Map Platform | Vague/Complex queries | Spatial hallucination |
| Jin et al. (2024b) | LLM | Visual context + Thinking prompts | Driving thought; Control action | Human-like driving simulation | Simulator (CARLA) | Explainability; Mechanical behavior | Data collection cost; Inference latency |
| Liu et al. (2023b) | LLM | List of locations; Semantic constraints | Optimized route sequence | Vehicle routing | Real-world data | Semantic constraints | Calculation accuracy |
| Wen et al. (2023) | LLM | Environment description | Action; Updated experience | Highway / Intersection | Simulator (Highway-env / CitySim) | Catastrophic forgetting; One-shot learning | Retrieval latency; Memory conflict |
| de Zarzà et al. (2023) | LLM | System error; State description | PID parameters | Truck platooning | Simulator (Network/Control sim) | Parameter tuning | Real-time latency; Safety stability |
| Sha et al. (2023) | LLM | Scene description | Cost function weights / constraints for MPC | Complex maneuvers | Simulator (IdSim / High-fidelity sim) | Quantifying abstract goals | Interface complexity; Inference speed |
| Mao et al. (2023b) | LLM | Perception results; Memory | Trajectory & Action explanation | Urban driving | Dataset (nuScenes) | Cognitive reasoning | Hallucination; Latency |
| Chen et al. (2024b) | LLM | Object-level vectors | Control action; Explanation | Urban driving | Real-world data (Wayve proprietary) | Modality mismatch | Dependency on perception |
| Hu et al. (2023a) | LLM | Video; Text; Action | Future video frames | General driving | Real-world data | Simulation realism | Hallucination; Compute intensive |
| Zhang et al. (2025b) | VLM | Images / BEV features | Lane topology graph | Online map construction | Dataset (OpenLane-V2 / nuScenes) | Topological errors | Inference speed |
| Liu et al. | VLM | Multi-view | Trajectory / | Complex | Simulator | Attention | Attention |
| (2025) | | images + Text commands | Control signals | urban driving | (CARLA) | distraction | labeling; Computational overhead |
| You et al. (2025) | VLM | Ego camera; Cooperative view; V2X | Trajectory / Safe control action | Cooperative driving | Simulation (Cooperative scenarios) | Single-vehicle blindness | Communication latency; Data sync |
| Zhang et al. (2024c) | VLM | Key-frame images | Driving action | Urban complex scenarios | Dataset (nuScenes) | Black-box decision | Error propagation; Inference speed |
| Zhao et al. (2024b) | VLM | Front-view images; Prompts | Behavior decision (text); Explanation | Urban / Highway | Dataset (Jimu / BDD100K based) | Decision interpretability | Open-loop; 2D-only |
| Yang et al. (2025b) | VLM | Multi-view images; Text prompts | Trajectory; Control signal; Explanation | End-to-end driving | S imulator (nuScenes / CARLA) | Computational efficiency | Training instability; Memory overhead |
| Yuan et al. (2024) | VLM | Current image | Natural language explanation | Driving explanation | Dataset (BDD-OIA / nuScenes) | Hallucination; Few-shot generalization | Retrieval latency |
| Zhang et al. (2025a) | VLM | Visual data; Safety rules | Safe trajectory | Safety-critical scenarios | Simulator (CARLA / ScenarioNet) | Safety assurance | Knowledge conflict; Annotation cost |
| Aghzal et al. (2025) | VLM | Scene image; Visualized path | Score | Path planning evaluation | Planning benchmarks | Metric design | Visual resolution |
| Long et al. (2024) | VLM | Camera images + Ego state | Low-level control | Complex maneuvers | Simulator | Dynamic feasibility | Guidance frequency |
| Tian et al. (2024) | VLM | Image sequence | Perception -> Analysis -> Planning | Urban complex & Long-tail scenarios | Dataset (DriveVLM-Dual / nuScenes) | Black-box reasoning; Slow-thinking capability | Inference latency; Real-time constraints |
| Renz et al. (2025) | VLA | Camera images + Language instructions | Control actions | Closed-loop urban driving | Simulator (CARLA) | Feature disconnect | Sim-to-real gap; Instruction granularity |
| Shao et al. (2024) | VLA | Multi-view images + Navigation instructions | Control actions; Predicted waypoints | Language-guided navigation | Simulator (CARLA) | Closed-loop evaluation | Latency; Dependence on teacher |
| Renz et al. (2024) | VLA | Monocular / Multi-view images | Actions | Camera-only closed-loop driving | Simulator (CARLA / Wayve Internal) | Model architecture | Tokenization efficiency |
| Jia et al. (2023) | VLA | Video history; Text instruction | Video prediction + Control actions | General driving / Predictive simulation | Dataset (nuScenes) | Unified representation | Inference speed |
| Chi et al. (2025) | VLA | Multi-view video; Language instructions | Control actions | General driving scenarios | Open datasets | Reproducibility & Accessibility | Compute requirement |
| Zhou et al. (2025) | VLA | Visual observation; Task prompt | Reasoning chain; Precision control | Closed-loop end-to-end driving | Simulator | Control precision | Training stability; Inference latency |

### *3.3.1. Foundational Generative Models*

The integration of foundational generative models marks a significant paradigm shift in AD motion planning transitioning from deterministic optimization to generative decision-making. Unlike traditional rule-based or purely predictive methods, generative models—specifically Variational Autoencoders (VAEs) (Kingma and Welling, 2013), Generative Adversarial Networks (GANs) (Goodfellow et al., 2020), and Diffusion Models (Ho et al., 2020)—excel at modeling the multimodality of human behavior and the complex joint distribution of future trajectories (Janner et al., 2022). This capability enables planners to sample diverse, kinematically feasible solutions, particularly in long-tail scenarios.

Early generative approaches focused on capturing the stochastic nature of traffic interactions. Early approaches utilized GANs to generate socially plausible trajectories via adversarial training aimed at mimicking human-like interactions (Gupta et al., 2018), however, these models frequently suffer from mode collapse. To enhance interpretability and stability, VAEs have been widely adopted to learn structured latent spaces of driving behaviors. For instance, Trajectron++ (Salzmann et al., 2020) utilizes a Conditional VAE (CVAE) framework to produce multimodal probabilistic predictions that inform downstream planning. Building on this, GenAD (Zheng et al., 2024a) extends the VAE paradigm to an end-to-end framework, modeling the future trajectory distribution in a structural latent space to ensure generated plans conform to real-world kinematic characteristics.

Recently, Diffusion Models have emerged as the state-of-the-art for motion planning, offering superior stability over GANs and broader mode coverage than VAEs. Inspired by the Diffuser framework (Jiang et al., 2023) which treats planning as a “probabilistic inpainting” problem, recent works apply iterative denoising to progressively refine trajectories from random noise under complex constraints. Gen-Drive (Huang et al., 2025c) exemplifies this “generation-evaluation” paradigm, combining a diffusion-based scenario generator with a Transformer-based evaluator to optimize decision-making in diverse scenarios. Similarly, approaches like SceneDiffuser (Jiang et al., 2024) demonstrate that diffusion policies can effectively handle long-horizon consistency and multi-agent interactions, addressing the limitations of single-step policies.

#### *3.3.2. LLM Based Methods*

LLMs, which have gained substantial attention for their capacity to emulate human reasoning, have introduced a novel paradigm for end-to-end AD. These models excel at understanding complex, unstructured environments and reasoning with common sense, addressing key limitations of traditional methods in handling long-tail scenarios and providing interpretability.

Owing to their distinctive capabilities, LLMs have been applied to every stage of the AD pipeline. In terms of perception and prediction, LLMs are employed to parse semantic information from heterogeneous sources (Fu et al., 2024)—ranging from LiDAR point clouds (Yang et al., 2025a) and BEV features (Tang et al., 2025) to pedestrian communications (Park et al., 2024)—and to detect semantic anomalies (Elhafsi et al., 2023). In prediction tasks, they utilize natural language descriptions to infer motion intentions and future trajectories (Chib and Singh, 2025; Keysan et al., 2023). Regarding planning and control, LLMs function as knowledge-driven engines for decision-making. These models can infer user intentions for route planning (Zhang et al., 2023a), learn strategies from expert demonstrations (Jin et al., 2024b; Liu et al., 2023b), and leverage memory modules for experience-based continuous evolution (Wen et al., 2023). However, due to strict real-time safety constraints, LLMs are rarely deployed as direct vehicle controllers (Zhu et al., 2025). Instead, a hierarchical framework is widely adopted, where the LLM acts as a high-level decision-maker while traditional controllers (e.g., MPC, PID) handle low-level execution (de Zarzà et al., 2023; Sha et al., 2023).

To mitigate the error accumulation and latency inherent in modular pipelines, recent research explores employing LLMs as central agents in end-to-end architectures. These systems process driving information holistically to generate human-like decisions (Mao et al., 2023b). Innovative approaches include fusing structured vector data to reduce training costs (Chen et al., 2024b) and developing generative world models to predict diverse outcomes for long-tail scenarios (Hu et al., 2023a).

Despite their reasoning capabilities standard LLMs are inherently limited by their text-only input modality, often requiring complex adapters to process visual data. This limitation has spurred the transition toward Multimodal Large Language Models (MLLMs), which natively integrate visual perception with linguistic reasoning.

#### *3.3.3. VLM Based Methods*

While LLMs demonstrate exceptional reasoning capabilities, their inability to process visual information restricts their direct application in dynamic traffic environments. VLMs address this gap by serving as a critical bridge between raw sensor data and high-level semantic reasoning. As illustrated in Fig. 10(a), VLMs ground visual inputs in linguistic semantics, thereby enhancing environmental perception, enabling interpretable decision-making, and improving robustness in rare "corner cases".

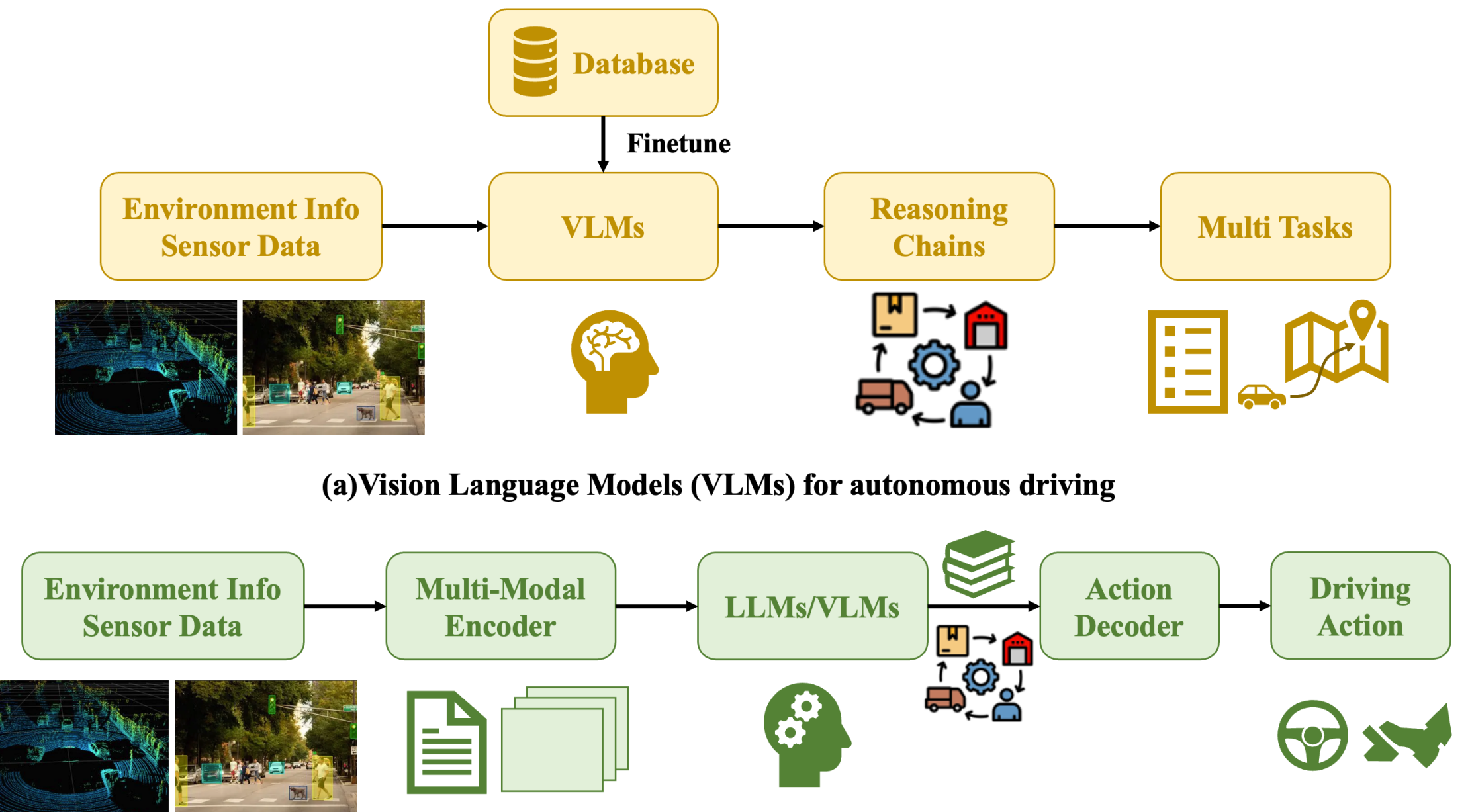

**Fig. 10.** Comparison of VLM and VLA for AD modes (Jiang et al., 2025). (a) VLMs for AD introduces natural language reasoning and explainability, yet remains perception-centric. (b) VLA for AD integrates perception, reasoning, and action, enabling interpretable and robust closed-loop control.

Regarding semantic perception and long-tail generalization, VLMs are characterized by their ability to interpret complex scenes using information beyond the geometric primitives typically employed in traditional pipelines. In the perception stage, VLMs are employed not just for object detection but for deep semantic understanding. For instance, Chameleon (Zhang et al., 2025b) leverages VLMs to infer complex lane topologies at urban intersections, providing crucial structural context for navigation. Similarly, VLM-E2E (Liu et al., 2025) fuses traditional BEV features with VLM-generated text annotations, yielding a more robust representation for downstream planning. Zhou et al. (Zhou et al., 2024) systematically validated these capabilities by evaluating VLMs on scenario-based questions (e.g., traffic signs and driving tests), confirming their potential as foundational components for situational awareness. Crucially, VLMs offer a promising solution to the "long-tail" problem, where data-driven models often fail due to the scarcity of training samples. Their vast, pre-trained knowledge bases enable reasoning about statistically underrepresented events. Benchmarks such as CODA-LM (Chen et al., 2025) have been developed to systematically evaluate capabilities in extreme scenarios. Building on this, CODA-VLM (Chen et al., 2025) has demonstrated state-of-the-art performance in identifying and reasoning about corner cases compared to open-source counterparts. Furthermore, generative frameworks like SEAL (You et al., 2025) utilize VLMs to synthetically generate adverse scenarios (e.g., dense fog, heavy snow), creating valuable training data to enhance the robustness of planning models.

Beyond perception, VLMs are increasingly integrated into the control architecture to provide interpretable, high-level driving decisions. Unlike "black-box" end-to-end models, VLM-based planners can articulate the rationale behind their actions. Frameworks like Think-Driver (Zhang et al., 2024c) exemplify this shift, moving from passive scene understanding to active decision-making with Chain-of-Thought (CoT) reasoning. Similarly, DriveLLaVA (Zhao et al., 2024b) generates interpretable driving behaviors by processing combined image and text inputs. Notably, it supports few-shot learning, allowing the system to adapt quickly to novel scenarios with more structured and refined behavioral descriptions than earlier LLM-based agents such as GPT-Driver (Mao et al., 2023a). Similarly, related works (Gao et al., 2025; Yang et al., 2025b; Yuan et al., 2024) have deeply integrated natural language into the planning and decision-making core of autonomous vehicles, leveraging language to understand context, coordinate vehicles, or generate driving behaviors, thereby bridging the semantic gap between high-level commands and low-level vehicle control.

Despite their semantic capabilities, deploying VLMs on actual vehicles remains challenging due to hallucinations (Aghzal et al., 2025) and high inference latency, which conflict with the strict safety requirements of traffic engineering. Consequently, the field has converged toward hybrid architectures that decouple high-level reasoning from low-level execution. Instead of replacing traditional controllers, these frameworks treat the VLM as a strategic guide, utilizing robust control algorithms (e.g., MPC) or parallel 3D planning stacks to ensure execution feasibility and spatial precision (Long et al., 2024; Tian et al., 2024). This hybrid paradigm successfully bridges the gap between open-loop reasoning and safety-critical deployment in mass-production vehicles.

In summary, VLMs enrich the AD pipeline by providing human-like understanding of complex scenarios. However, the reliance on hybrid architectures highlights the difficulty of achieving unified perception-action loops with VLMs alone. This limitation has paved the way for Vision-Language-Action models (VLAs), which aim to unify perception and control into a single end-to-end policy, as discussed in the following section

*3.3.4. VLA Based Methods*

VLA models represent a further evolution, functioning as unified agents that directly connect multimodal perception and reasoning to physical actions. Unlike VLMs that primarily focus on understanding and description, VLAs are designed to generate executable control commands or motion plans, thus embodying a true end-to-end approach. A primary objective of VLAs is to mitigate the "black-box" nature of end-to-end networks by rendering decisions and resulting actions explainable through natural language.

The core architecture of a VLA model for AD (Fig. 10(b)) typically consists of three key components: a multimodal encoder, an LLM/VLM processor, and an action decoder. The system receives environmental data from multi-modal sensors like cameras and LiDAR, along with natural language commands, as input. A vision encoder first converts visual information into latent representations, which are then fused with language instructions inside the LLM/VLM core for reasoning. Finally, an action decoder translates the model's reasoning into concrete driving actions, such as generating an executable trajectory plan, predicting waypoints, or directly outputting low-level control signals for steering, throttle, and braking.

According to Jiang et al. (Jiang et al., 2025), the evolution of VLA models can be categorized into four stages: LLM as explainer, modular VLA, unified end-to-end VLA, and reasoning-augmented VLA. Given that the first two stages—which utilize models for scene description or modular VLA—share significant methodological overlaps with the approaches detailed in Sections 3.3.2 and 3.3.3, this section focuses on the latter stages that represent the critical leap toward unified, embodied intelligence.

To address the latency and error cascading inherent in modular pipelines, recent research has focused on unified end-to-end architectures that integrate perception, reasoning, and action into a single differentiable system. SimLingo (Renz et al., 2025) exemplifies this by aligning language understanding with "Action Dreaming" technology for controllable driving. Similarly, LMDrive (Shao et al., 2024) and CarLLaVA (Renz et al., 2024) map sensor inputs directly to control signals. Furthermore, ADriver-I (Jia et al., 2023) extends this concept by incorporating diffusion-based world models to predict future frames based on action hypotheses, enabling foresighted planning based on generative predictions. In recent years, VLA development has entered a new stage, with research shifting toward long-term reasoning, memory, and interaction, placing the LLM/VLM at the center of the control loop. Impromptu VLA (Chi et al., 2025) explicitly links CoT reasoning with action generation, ensuring the model "thinks" before it acts to improve decision accuracy in unstructured environments. Concurrently, AutoVLA (Zhou et al., 2025) unifies reasoning and action generation in a single autoregressive model, significantly improving decision interpretability and generalization in complex scenarios through adaptive reasoning and reinforcement fine-tuning.

Despite significant progress, VLAs face critical challenges. Ensuring the physical plausibility and safety of generated actions remains a primary concern, as a misinterpretation of language or visual input could lead to dangerous maneuvers. The real-time performance of these large models is another hurdle for onboard deployment. Future work will likely focus on developing more efficient VLA architectures, improving their reasoning capabilities in dynamic and interactive environments, and expanding datasets to cover an even wider range of complex driving situations.

### 3.4. Challenges of Learning-Based Motion Planning

The preceding sections have detailed the significant advancements brought by IL, RL, and Generative AI to the field of motion planning. These learning-based methods have demonstrated a remarkable ability to handle complex, unstructured driving environments by learning directly from data, thereby addressing many of the limitations of traditional, rule-based pipeline systems. However, despite their promise, these end-to-end or purely learning-driven paradigms share a set of fundamental challenges that hinder their widespread adoption in safety-critical real-world applications.

A primary and intrinsic concern lies in the lack of verifiable safety and interpretability. Most learning-based systems—from IL to advanced MLLMs—operate as “black boxes”, rendering their decision-making processes opaque. Although MLLMs can provide natural language rationales for their outputs, such explanations do not fundamentally resolve the inaccessibility of their internal logic. For end-to-end systems, this opacity poses serious difficulties for debugging and formal verification. Beyond interpretability, the trial-and-error nature of RL complicates safety guarantees, as agents will explore unsafe behaviors during training and inference processes. More critically, models such as VLMs and VLAs are known to exhibit “hallucinations”, misinterpreting scenes and generating potentially catastrophic actions. This inherent opaqueness remains a major barrier to public trust and regulatory approval.

A second challenge is the pronounced inefficiency in terms of data and computation. IL requires vast quantities of curated, high-quality expert demonstrations, which are costly and difficult to obtain. However, RL suffers from low data efficiency, typically requiring extensive interaction to achieve convergence. Meanwhile, despite their powerful capabilities, modern Generative AI models impose enormous computational burdens. Large-scale LLMs, VLMs, and VLAs present severe obstacles to achieving real-time performance on resource-constrained onboard hardware.

Finally, a higher-level expectation for AD is the attainment of human-like intelligence. As discussed in Section 2.2, such systems should demonstrate capabilities of evolution, personalization, and interactive driving. Yet, current learning-based methods struggle to meet these requirements. Even state-of-the-art MLLMs, despite their impressive reasoning abilities, fall short of delivering human-satisfactory levels of anthropomorphic driving behavior.

Taken together, these challenges underscore the necessity of a new paradigm. The fundamental issues: (1) the absence of verifiable safety and interpretability, (2) severe bottlenecks in data and computational efficiency, and (3) difficulties in achieving human-like intelligence, suggest that relying solely on learning-based methods may not constitute the optimal path forward. This motivates the investigation of hybrid architectures that combine the adaptability and perception strength of machine learning with the rigor, structure, and theoretical foundations of classical control. Such architectures are designed to preserve the advantages of data-driven learning while embedding explicit constraints, parameterization, and optimality in a principled manner. This perspective naturally leads to the study of data-driven optimal control.

## 4. Data-Driven Optimal Control Approaches

The analysis in Sections 2 and 3 identifies a dichotomy in current motion planning methodologies. While pipeline methods offer theoretical interpretability and modularity, they are prone to local optimality and structural brittleness in complex scenarios. Conversely, purely learning-based approaches (e.g., IL, RL,

Generative AI) demonstrate superior adaptability to complex scenarios but face significant challenges regarding interpretability, safety guarantees, and data efficiency. To address these limitations, this paper advocates for a DDOC paradigm. This approach aims to synergize the rigorous structural guarantees of classical control with the adaptive capabilities of modern machine learning.

While Optimal Control theory encompasses a broad spectrum of mathematical optimization techniques, MPC has emerged as the most dominant and effective realization in the context of AD motion planning due to its ability to explicitly handle constraints and future horizons. Consequently, strictly speaking, the implementation of DDOC in this field largely manifests as Data-Driven Predictive Control (DDPC). This paradigm is founded on MPC. Unlike black-box policies, MPC explicitly formulates motion planning as a constrained optimization problem, offering a theoretical framework to enforce safety constraints and kinematic feasibility. The standard mathematical formulation is expressed as:

$$\begin{aligned} &\min_{\boldsymbol{u}_k,\ldots,\boldsymbol{u}_{k+N-1}} \sum_{i=k}^{k+N-1} J(\boldsymbol{x}_i,\boldsymbol{u}_i) \\ &\text{s.t. } \boldsymbol{x}_{i+1} = f(\boldsymbol{x}_i,\boldsymbol{u}_i) \\ &\quad \boldsymbol{x}_i \in \boldsymbol{\mathcal{X}} \\ &\quad \boldsymbol{u}_i \in \boldsymbol{\mathcal{U}} \\ &\quad i = k,\ldots,k+N-1 \end{aligned} \tag{6}$$

where $\boldsymbol{\mathcal{X}}$ and $\boldsymbol{\mathcal{U}}$ denote the feasible regions of state and control respectively; $k$ represents the current time step. Traditional MPC relies on accurate vehicle dynamics models and environmental data. However, in real-world scenarios, road conditions are highly complex and variable, and vehicle dynamics parameters can change over time, making precise modeling challenging.

To address these modeling limitations, researchers have turned to Data-Driven Control (DDC). To clarify the terminology used in this section, we provide a compact taxonomy of DDC methods, as illustrated in **Fig. 11**. DDC is a comprehensive category for methodologies that design control systems using online or offline input–output data to satisfy key control objectives. When DDC principles are integrated into the MPC framework to enhance the accuracy of predictions or the adaptability of objectives, the resulting methodology is termed DDPC. Leveraging online learning, adaptive data modeling, and model updating, DDPC addresses the shortcomings of traditional MPC (Rosolia et al., 2018), demonstrating superior capabilities in customization, dynamics adaptation, and self-tuning.

As illustrated in the literature (Krishnan and Pasqualetti, 2021), DDPC approaches can be categorized into indirect DDPC (e.g., Learning-Based Model Predictive Control, LBMPC) and direct DDPC (e.g., Data-enabled Predictive Control, DeePC), based on how they utilize data. Indirect DDPC follows a sequential "Identify-then-Control" paradigm. Here, system identification techniques—such as Subspace Predictive Control (SPC) or neural network-based approaches—are employed to construct an explicit model that is subsequently incorporated into MPC constraints. While this approach benefits from model-based theoretical guarantees, it is often sensitive to model structure selection and suffers from a fundamental "objective mismatch," where the identification phase prioritizes prediction accuracy rather than the ultimate control performance. Conversely, Direct DDPC eliminates the explicit identification bottleneck by leveraging raw or pre-processed data to predict the system's future response directly. Grounded in theories such as Willems' Fundamental Lemma, these methods bypass the construction of a parametric model, thereby allowing the solver to map available data directly to control objectives and reducing the risk of modeling errors. However, practical application in AD is often constrained by high computational costs associated with large datasets and inherent sensitivity to noise in nonlinear environments.

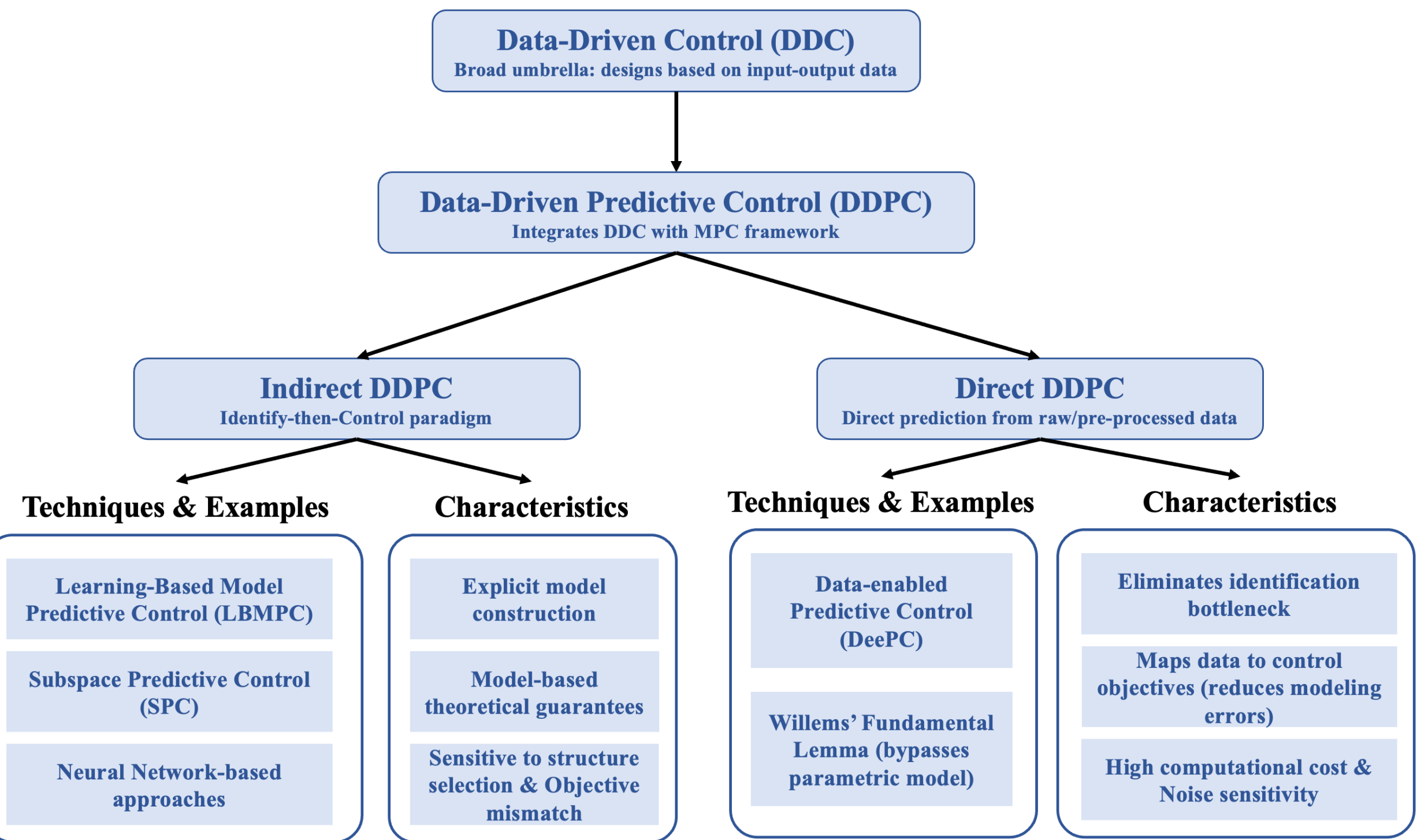

**Fig. 11.** Taxonomy of DDC Methods

This section examines the pivotal role of data-driven MPC in AD motion planning, arguing that DDOC provides a robust pathway toward trustworthy autonomy. **Table 8** summarizes all prominent DDPC methods reviewed in this part. The remainder of this section is organized as follows: Section 4.1 discusses the technical mechanisms by which learning algorithms empower DDPC. Sections 4.2 through 4.4 explore the three critical capabilities enabled by this paradigm: customization, dynamics adaptation, and self-tuning. Finally, Section 4.5 addresses the current limitations and open challenges of DDPC methods.

**Table 8** A summary of representative DDPC-based motion planning methods for AD

| Article | Method type | Customization | Dynamic Adaptation | Self-Tuning |
|---|---|---|---|---|
| Wang et al. (2021) | Indirect DDPC | | ✓ | |
| Zhang et al. (2021a) | Indirect DDPC | | ✓ | |
| Rokonuzzaman et al. (2021) | Indirect DDPC | | ✓ | |
| Edwards et al. (2021) | Indirect DDPC | | | ✓ |
| Jafarzadeh and Fleming (2021) | Indirect DDPC | | | ✓ |
| Rokonuzzaman et al. (2022) | Indirect DDPC | ✓ | | |
| Ren and Xi (2022) | Indirect DDPC | | ✓ | |
| Wang et al. (2022d) | Direct DDPC | | | ✓ |
| Zhang et al. (2023c) | Indirect DDPC | ✓ | | |
| Wang et al. (2023) | Indirect DDPC | ✓ | | |
| Sha et al. (2023) | Indirect DDPC | ✓ | | |
| Hu et al. (2025a) | Indirect DDPC | ✓ | | |
| Picotti (2024) | Indirect DDPC | | ✓ | |
| Shaiakhmetov et al. (2024) | Direct DDPC | | ✓ | |
| Zhang et al. (2024a) | Direct DDPC | | | ✓ |
| He et al. (2025) | Direct DDPC | ✓ | | |
| Lian et al. (2025) | Indirect DDPC | | | ✓ |
| Hu et al. (2025b) | Indirect DDPC | | | ✓ |

## 4.1. Learning-Empowered Optimal Control Formulation

The analysis in the preceding sections highlights a fundamental dichotomy in motion planning: purely learning-based methods offer superior adaptability to complex environments but often lack interpretability and verifiable safety, whereas traditional pipeline methods provide rigorous structural guarantees but suffer from brittleness and local optimality. To resolve this impasse, DDOC does not discard the theoretical foundations of classical control; rather, it adopts a hybrid architecture that synergizes the strengths of both paradigms. In this framework, the MPC formulation (Eq. (6)) serves as the verifiable backbone, strictly enforcing safety constraints and kinematic feasibility through optimization. Concurrently, the learning algorithms reviewed in Section 3—ranging from IL to Generative AI—are integrated as intelligent "estimators" or "generators" that parameterize the internal components of the control loop. This paradigm shift transforms traditional white-box MPC into a learning-empowered gray-box system, specifically enhancing the objective function, system dynamics, and optimization policy.

The first critical integration targets the cost function. In traditional control, constructing a cost function that accurately balances safety, efficiency, and comfort is a manual, trial-and-error process that fails to quantify complex human intentions. To address this, multiple learning paradigms are integrated into the formulation. IRL allows the system to implicitly learn the structure and weights of the cost function directly from expert demonstration data, capturing nuanced driving styles. Simultaneously, LLMs and VLMs can function as semantic interpreters, translating unstructured natural language instructions or scene contexts into structured cost constraints. Furthermore, models trained via RL can also dynamically adjust these weights based on real-time feedback. Collectively, these methods transform the cost function from a fixed equation into an adaptive interface, providing the mathematical foundation for both Customization (Section 4.2) and Self-Tuning (Section 4.4).

The second critical integration targets the system dynamics. Traditional MPC relies on simplified kinematic models (e.g., bicycle models), which often fail to account for highly nonlinear vehicle dynamics, mechanical wear, or complex environmental interactions. Learning algorithms upgrade this component from idealized physics to high-fidelity data-driven representations. SL techniques, such as Neural Networks (NNs) and Gaussian Processes (GPs), are employed to model the residual errors of physical dynamics or to learn the vehicle's response function from scratch, compensating for internal variations. Expanding the scope, generative world models and Diffusion Models are integrated to predict the probabilistic evolution of the surrounding environment and agents. By enabling the controller to anticipate both vehicle-level and scene-level changes, these methods underpin the Dynamics Adaptation (Section 4.3) while also enhancing the precision required for high-performance Self-Tuning (Section 4.4).

Finally, learning algorithms empower the optimization policy and constraints, shifting from static parameter settings to intelligent self-tuning. The MPC solving process is traditionally governed by fixed hyperparameters (e.g., prediction horizon $N$) and static constraints, limiting adaptability and computational efficiency in dynamic scenarios. In the DDOC framework, RL agents are deployed as meta-controllers to dynamically tune these hyperparameters in real-time, balancing the trade-off between foresight and computational cost. Additionally, IL can provide high-quality initial guesses (warm starts) to accelerate the optimization solver, avoiding local optima. By making the solving process itself intelligent and enabling hyperparameters to adapt automatically, these techniques directly enable Self-Tuning (Section 4.4) and allow for the dynamic adjustment of safety margins to support diverse driving preferences (Section 4.2).

By embedding these learning-based components into the optimal control backbone, DDOC achieves a synergy where learning provides adaptability and control theory ensures safety and feasibility.

### 4.2. Data-Driven for Customization

Customization is a core challenge in enhancing the human-machine interaction and acceptance of AD systems. In traditional MPC frameworks, the parameters—specifically the weights in the objective function and the boundaries of safety constraints—are typically set statically based on an engineer's expertise. This "one-size-fits-all" approach fails to adapt to the personalized preferences of different users, such as varying demands for acceleration responsiveness in eco-driving or diverse risk tolerances in lane-changing scenarios. Data-driven MPC addresses this by establishing a mapping between driver operation data (e.g., pedal travel, steering rates) and control parameters, enabling a highly personalized planning paradigm. Unlike the black-box nature of end-to-end learning methods, data-driven customization maintains the rigorous safety structure of control theory while imparting the system with human-like adaptability. Recent advancements in this domain have evolved from simple offline parameter fitting to sophisticated online inference and semantic alignment, focusing on three key dimensions: objective function learning, dynamic constraint adaptation, and semantic-to-parameter mapping.

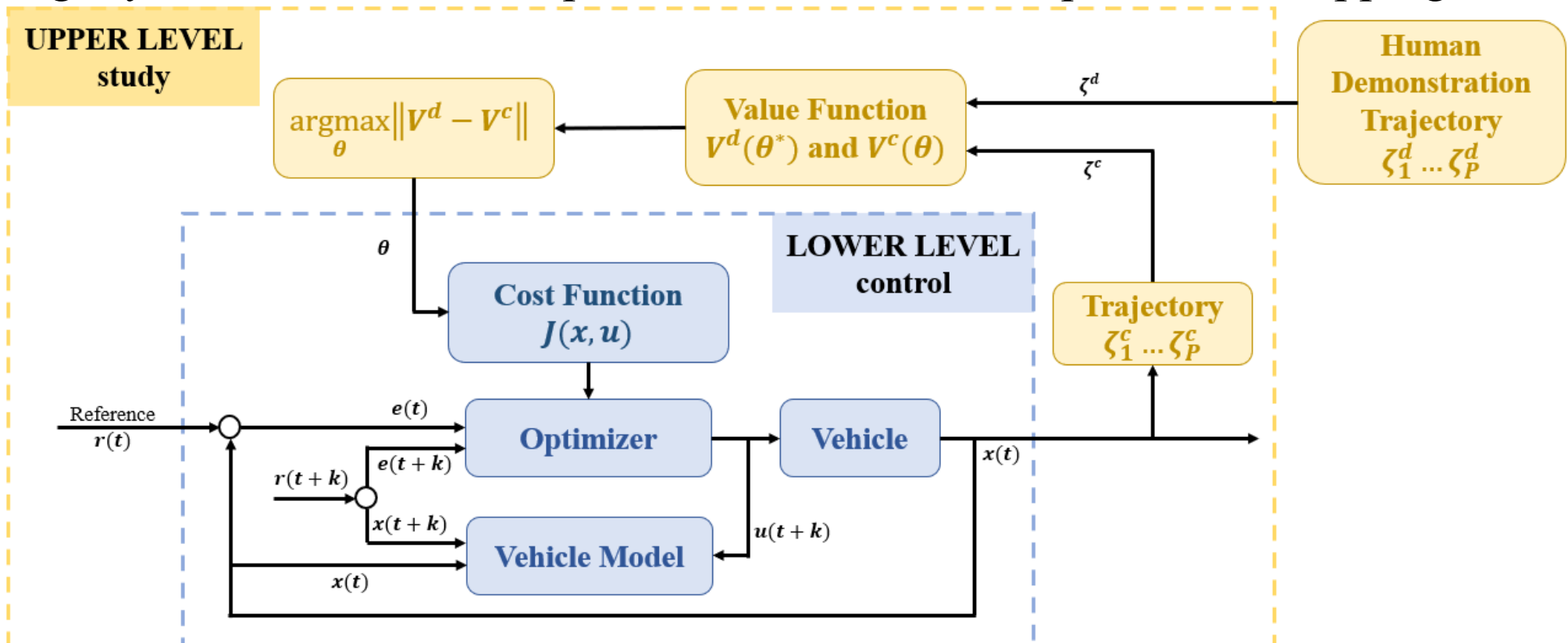

**Fig. 12.** An IOC method based on bi-level optimization for learning cost function parameters within an MPC framework (Rokonuzzaman et al., 2022).

The most direct customization method is to infer the driver's internal utility function. Instead of manually tuning cost function parameters in Eq. (6), recent works utilize Inverse Optimal Control (IOC) or IRL to learn these weights from demonstration data. Rokonuzzaman et al. (Rokonuzzaman et al., 2022) proposed a LBMPC method for personalized AD trajectory planning (**Fig. 12**). They designed a combined lateral and longitudinal MPC controller with a feature-based parametric cost function. This approach utilizes inverse optimal control to learn MPC cost function parameters from human personalized driving data, effectively regenerating individual driving characteristics. Moving beyond offline learning, Zhang et al. (Zhang et al., 2023c) introduced an integrated behavior-motion planner capable of Online IRL. In complex interactive scenarios like unsignalized intersections, this system infers the personalized cost function of the ego vehicle in real-time based on the aggressiveness of surrounding agents, allowing the AD system to dynamically adjust its behavior to secure right-of-way. Similarly, the Anti-bullying ACC (Hu et al., 2024a) utilizes online IOC to identify the intruder's driving style and employs a game-theoretic MPC to proactively protect the right-of-way.

Customization extends beyond comfort (cost function) to risk tolerance (constraints). Different drivers perceive "safety" differently, necessitating adaptive constraint boundaries. In this context, recent advancements in planning methodologies provide the theoretical basis for constructing Data-Driven MPC Constraints. For instance, the "Dynamic Safety Domain" concept proposed by Wang et al. (Wang et al., 2023) serves as a blueprint for formulating time-varying state constraints. By mapping driver behavior to the geometric parameters of the safety envelope, this approach enables the MPC solver to generate maneuvers aligning with human spatial intuition rather than rigid, conservative buffers. Similarly, the "Lesson Learning" framework was developed by Hu et al. (Hu et al., 2025a). offers a mechanism for the online refinement of these constraints. By treating human takeover events as explicit boundary violation signals, this logic can be adapted into the DDPC framework to iteratively tighten or loosen safety constraints, effectively evolving the controller's operational envelope to match the user's specific comfort zone.

Due to the powerful semantic understanding ability of LLMs, an emerging trend involves bridging the gap between high-level user intent and low-level control parameters using LLMs. Sha et al. (Sha et al., 2023) proposed LanguageMPC, which utilizes LLMs as a reasoning engine to process natural language instructions (e.g., "I am in a hurry") and scene descriptions, outputting specific weight coefficients for the MPC cost function. This approach provides an interpretable interface for customization, allowing users to define the vehicle's driving style through natural conversation rather than engineering tuning.

In summary, data-driven customized MPC has emerged as a pivotal approach for achieving personalized AD. By leveraging techniques ranging from bi-level optimization and online IRL to LLM-based reasoning, these methods successfully quantify user driving styles into interpretable MPC parameters, such as weights and constraints. This paradigm preserves the theoretical safety boundaries of optimal control while endowing the system with the ability to continuously evolve and align with diverse human behaviors.

### 4.3. Data-Driven for Dynamics Adaptation

Autonomous vehicles must contend with highly time-varying dynamic characteristics. The same vehicle model, for instance, might exhibit variations in acceleration response due to mechanical wear or load changes. More pronounced are the discrepancies in chassis dynamics across different vehicle platforms, such as SUVs and sedans. Traditional MPC, relying on fixed-parameter models, is prone to error accumulation and potential instability when facing such uncertainties. Data-driven dynamic adaptation mechanisms address this challenge by incorporating online learning modules. These modules continuously integrate real-time vehicle information, like actual wheel speed and yaw rate, with data from measurement units to update the system's dynamic equations in real time. This adaptive capability allows MPC to accommodate heterogeneous vehicle platforms without requiring precise prior models, significantly enhancing commercial deployment flexibility.

To mitigate the reliance on accurate physical priors, substantial research focuses on addressing the uncertainties during state estimation. Instead of discarding the nominal physical model, these approaches utilize data-driven methods to learn the "mismatch error" between the model and the actual system. Zhang et al. (Zhang et al., 2021a) applied this concept to the complex energy management of Plug-in Hybrid Electric Vehicles (PHEVs), integrating a GP model into the MPC framework. The GP learns the uncertainty in the state estimation process online, effectively optimizing energy flow under highly nonlinear constraints. Similarly, Picotti et al. (Picotti, 2024) validated the efficacy of GP-based MPC in vehicle dynamics control. Building on this philosophy, Ren et al. (Ren and Xi, 2022) proposed using Recurrent Neural Network (RNN) methods to model vehicle dynamic deviations. Notably, their research demonstrated that RNNs outperform classic GP models, particularly in capturing complex temporal dependencies within the dynamic mismatch. By quantifying the uncertainty of the learned model, these approaches enable the MPC controller to explicitly account for model inaccuracies, thereby improving robustness and ensuring constraint satisfaction under modeling errors and disturbances.

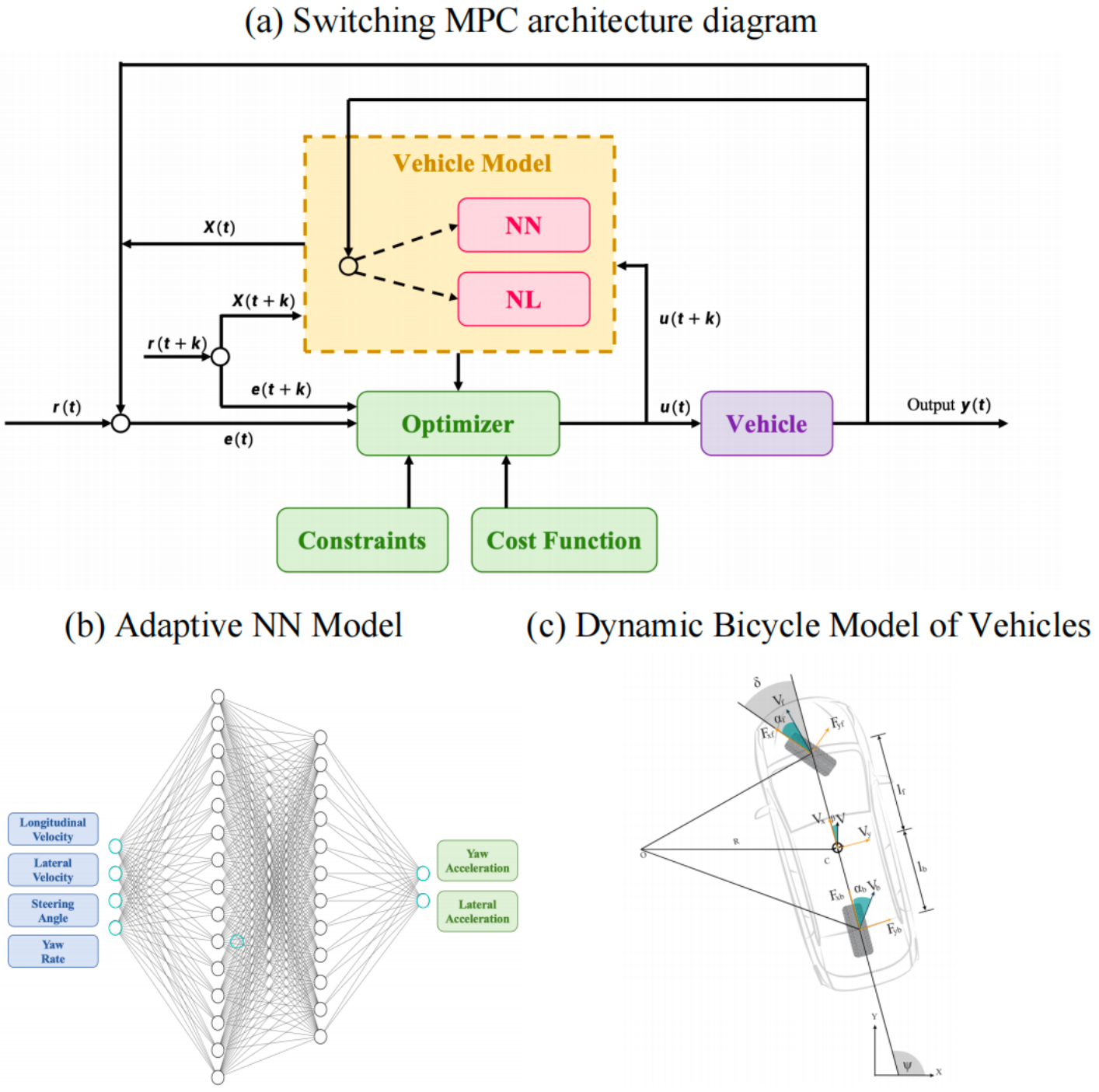

**Fig. 13.** The switchable MPC framework includes both adaptive NN models and traditional nonlinear physical models (Rokonuzzaman et al., 2021). In the early stages of operation or when facing new working

conditions, due to insufficient data, NN models often perform worse than nonlinear physical models. After learning real-time data and adapting to the environment, NN models often demonstrate better performance.

However, when facing strong non-linearities where nominal models fail, researchers turn to deep dynamics identification to reconstruct the system dynamics. Wang et al. (Wang et al., 2021) proposed a data-driven MPC design employing DNNs as Koopman operators for vehicle system identification. This method emphasizes learning dynamic evolution rather than explicitly modeling it, enabling vehicle dynamics to actively update based on recently collected data, thus allowing real-time adjustments. To address the “cold start” problem inherent in pure NNs, Rokonuzzaman et al. (Rokonuzzaman et al., 2021) proposed a data-driven switching MPC (**Fig. 13**) that leverages NNs to learn vehicle dynamics from the rich operational data of modern vehicle systems, demonstrating performance superior to traditional MPC.

In summary, by utilizing the data-driven MPC method, vehicles can perceive highly time-varying dynamic characteristics online (such as dynamic drift caused by wear, load, or platform differences), thereby upgrading the traditional paradigm of “accurate modeling first, robust control later” to a closed-loop architecture of “learning, adapting, and operating simultaneously”. This method not only significantly reduces the dependence on prior models but also provides a unified framework for rapid deployment across different vehicle models and operating conditions.

### 4.4. Data-Driven for Self-Tuning

While the structure of MPC provides safety guarantees, its real-world performance—defined by computational efficiency and adaptability—is constrained by its static configuration. Traditional MPC struggles with the constraints: complex models compromise real-time performance, while simplified models fail in dynamic scenarios. Data-driven self-tuning frameworks overcome this bottleneck by transforming the optimization process from a rigid execution of fixed rules into an adaptive, self-evolving system. This evolution manifests in optimized hyperparameters, enhanced model representation efficiency, and refined interaction dynamics.

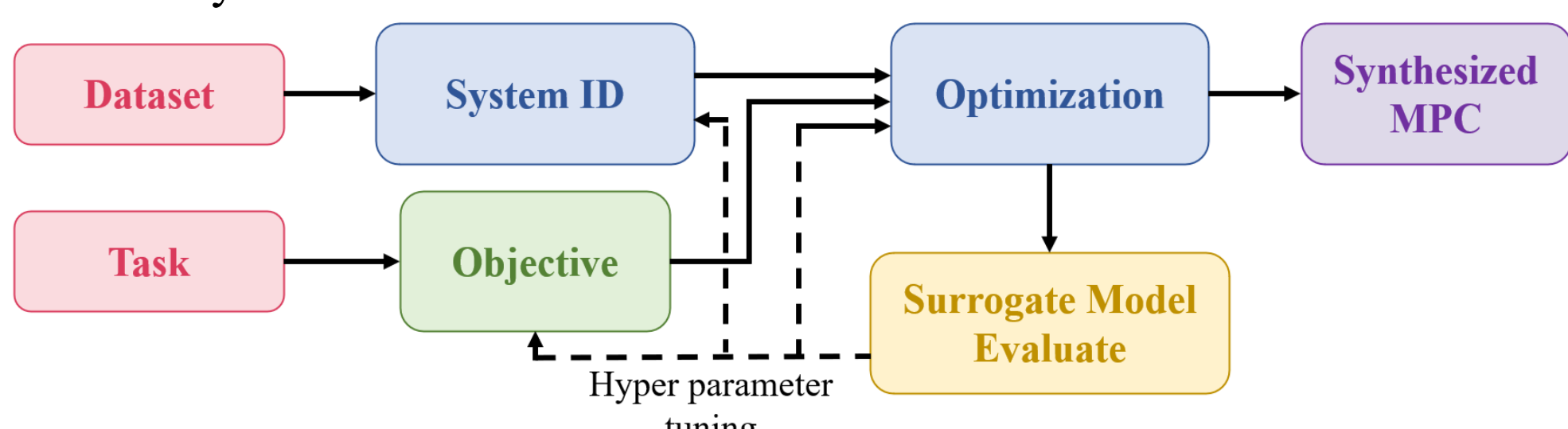

**Fig. 14.** AutoMPC software framework diagram (Edwards et al., 2021). In this framework, the user provides a Dataset containing system states and control trajectories, along with a specific Task. This task is then quantified as an Objective. The System ID module uses the input dataset to learn or identify a system dynamic model. Optimization leverages the System ID and Objective to solve for the optimal control sequence. The Surrogate Model Evaluate module performs closed-loop performance assessment, and Synthesized represents the final generated MPC controller.

The performance of MPC heavily relies on the selection of hyperparameters, such as the prediction horizon and cost function weights. However, traditional trial-and-error tuning struggles to meet the real-time demands of dynamic scenarios. For instance, a longer prediction horizon might improve lane change smoothness on highways but could lead to an infeasible optimization problem in dense traffic. Data-driven frameworks address this by constructing an implicit mapping between parameters and system performance. Edwards et al. (Edwards et al., 2021) proposed AutoMPC (**Fig. 14**), a data-driven framework capable of automatic optimization and adjustment. This framework has demonstrated robust performance across multiple tasks, and its efficiency and operational simplicity make it applicable to AD motion planning. Beyond parameters, defining tasks using explicit objective functions is challenging for complex behaviors. Jafarzadeh et al. (Jafarzadeh and Fleming, 2021) proposed a hybrid approach merging model-based and data-driven methods, formulated as:

$$
\begin{aligned}
J = & \min_{\boldsymbol{u}_k,\ldots,\boldsymbol{u}_{k+N-1}} \sum_{i=k}^{k+N-1} \left[h(\boldsymbol{x}_i,\boldsymbol{u}_i) + \hat{z}(\boldsymbol{x}_i,\boldsymbol{u}_i)\right] \\
\text{s.t. } & \boldsymbol{x}_{i+1} = f(\boldsymbol{x}_i,\boldsymbol{u}_i) \\
& \boldsymbol{x}_k = \boldsymbol{x}_s \\
& \boldsymbol{x}_i \in \boldsymbol{\mathcal{X}} \\
& \boldsymbol{u}_i \in \boldsymbol{\mathcal{U}} \\
& i = k,\ldots,k+N-1
\end{aligned} \tag{7}
$$

where $\boldsymbol{x}_s$ denotes the given initial system state at the start of the optimization, $h(\cdot)$ is a known model-driven function, and $\hat{z}(\cdot)$ is a data-driven term learned via machine learning to predict unknown cost components (e.g., hazardous areas or socially compliant behaviors), thereby enabling the MPC to optimize against complex, unmodeled environmental factors.

Achieving precise modeling that balances accuracy with computational efficiency is another hurdle. To address the computational latency of complex models, Zhang et al. (Zhang et al., 2024a) introduced a DDMPC method. Its core mechanism involves using a Hankel matrix, constructed from offline input-output trajectory data, to replace complex system dynamics. This enables the model to predict future states and calculate optimal inputs with reduced tracking error and computational time compared to traditional PID and kinematic-based MPC. Addressing the model mismatch issues of precise vehicle modeling, Kabzan et al. (Kabzan et al., 2019) employed an online learning controller utilizing Gaussian process regression. By accounting for model residual uncertainty, this method facilitates safe driving behavior near physical limits; testing demonstrated a 10% increase in vehicle speed within the safety envelope.

Another substantial limitation of traditional MPC is its simplified treatment of surrounding agents, which often fails to capture complex social interactions. Recent progress has focused on using the transformer architecture to model generalized system dynamics for vehicle interaction. Lian et al. (Lian et al., 2025) proposed ExpliDrive, a framework that explicitly bridges MPC and Transformers for interactive driving. By utilizing a Transformer-based encoder-decoder to capture multimodal interaction mechanisms and feeding these predictions into the MPC, the system effectively handles dynamic game-theoretic scenarios that traditional kinematic predictions cannot resolve. Building on this physics-informed direction, Hu et al. (Hu et al., 2025b) introduced MPCFormer. This approach formulates social interaction dynamics into a discrete state-space representation, with coefficients learned from naturalistic driving data via Transformers. By embedding these learned social dynamics into MPC constraints, the planner balances data-driven social awareness with physics-based explainability, significantly improving performance in dense, interactive traffic. Similarly, CogDrive (Huang et al., 2025a) addresses multimodal uncertainty by fusing topological interaction reasoning with contingency-based optimization.

In summary, the data-driven self-tuning framework fundamentally changes the paradigm of MPC from “offline fixation” to “online optimization.” By automating parameter tuning, employing efficient data structures for acceleration, and refining interaction models with Transformers, this paradigm significantly reduces the reliance on expert experience and provides a scalable solution for high-performance AD.

### 4.5. Limitations and Critical Challenges: The Reality Gap

While DDPC offers a promising synthesis of control-theoretic structure and learning-based adaptability, its deployment in real-world AD is non-trivial. This section explicitly discusses the failure modes and open challenges of DDPC, ranging from data paradoxes in safety regimes to instability under unmodeled dynamics and computational intractability.

#### *4.5.1. Data Requirements and the “Excitation” Paradox*

A fundamental premise of DDPC—particularly Direct DDPC methods relying on Willems’ Fundamental Lemma—is that the collected data must be sufficiently rich to characterize the system. However, this creates a conflict in safety-critical regimes, as theoretical guarantees often require input data to satisfy the condition of Persistent Excitation (PE). Recent theoretical insights by Shakouri et al. (Shakouri et al., 2025) reveal that PE is a necessary condition for universality; without it, the resulting input-output data cannot parametrize all finite-length trajectories. In AD, routine safe behavior does not sufficiently excite the vehicle dynamics, leading to data that is insufficient for identifying or learning critical system dynamics. Furthermore, even with sufficient data, the **generalization gap** remains a hurdle. As noted by Wu et al. (Wu et al., 2021b), the question regarding the generalization accuracy of learned models on Out-of-Distribution data has not been fully addressed. When the vehicle encounters a domain shift (e.g., unexpected weather or road friction changes), the "robustness" learned from offline data is often only limited to local regions (Vahidi-Moghaddam et al., 2025). This inability to handle new system dynamics that emerge online poses a severe risk to safety.

*4.5.2. Wrong Model Updates and Unmodeled Dynamics*

The adaptability of DDPC allows for online updates, but this plasticity introduces the risk of erroneous model updates and instability due to model-plant mismatch. Many direct data-driven methods assume a Linear Time-Invariant (LTI) structure. Vahidi-Moghaddam et al. (Vahidi-Moghaddam et al., 2025) point out that for nonlinear, stochastic, and time-varying vehicle systems, the standard rank condition of the Hankel matrix is often insufficient. Relying on such assumptions can lead to poor control performance or instability when the implicit linear model fails to capture the nonlinear reality of vehicle dynamics. Real-world sensor data is inevitably noisy. Coulson et al. (Coulson et al., 2019) highlight that algorithms using noise-corrupted I/O data for prediction require complex regularization or distributionally robust optimization to prevent divergence. Without these safeguards, the optimizer may interpret measurement noise as physical dynamics, leading to erratic control actions.

*4.5.3. Computational Burden and Constraint Ambiguity*

Finally, the engineering deployment of DDPC is constrained by computational burdens and the difficulty of tuning constraint sets. To mitigate the risks of model mismatch, DeePC often requires a large horizon of historical data. However, Li et al. (Li and Li, 2025) identify a critical bottleneck: the computational complexity of these methods often scales cubically with the dataset size. This creates a conflict between performance and real-time feasibility, often making standard DDPC intractable for onboard hardware without significant approximation. Moreover, while DDPC can learn cost functions and constraint parameters, distinguishing between "soft" and "hard" constraints from data remains an open challenge. Human drivers often treat constraints flexibly (e.g., crossing a line to avoid an accident). If a DDPC system learns these flexible boundaries as hard constraints, it may result in infeasibility; if interpreted as soft penalties, it may fail to guarantee safety. Therefore, automatically adjusting these constraint sets to balance safety and human-like flexibility is equally challenging.

## 5. Future directions

This paper examines the evolution of AD technology by analyzing traditional motion planning pipelines, prevalent learning-based end-to-end approaches, and data-driven MPC methods. While traditional pipelines struggle to guarantee global optimality, end-to-end AI methods inherently face challenges regarding safety and interpretability. DDOC offers a viable framework for integrating theoretical optimality with real-world adaptability. Building on this foundation, future research should evolve towards the following deeply integrated directions to construct a safer, more intelligent, efficient, and trustworthy motion planning system. **Table 9** provides a concise summary of these directions, outlining their typical structures, unresolved problems, and potential applications.

**Table 9** Summary of future directions

| Future directions | Typical structure | Open questions | Near-term applications |
|---|---|---|---|
| World Model-Centric MPC | **Generative Simulation:**<br>World models as internal predictors within the MPC loop. | 1. Closed-loop stability<br>2. Real-time feasibility | 1. Offline closed-loop validation<br>2. Low-speed logistics & shuttles |
| Hybrid Learning for Customized Control | **Bi-level Framework:**<br>Upper-level (LLM/IRL) tunes parameters for lower-level MPC. | 1. Balancing personalization vs. safety | 1. L2+ Navigate on Autopilot (NOA)<br>2. Personalized Highway Pilot |
| | | 2. Few-shot adaptation efficiency | |
| RL-Based Meta-Control | **Hierarchical Tuning:** RL agent dynamically optimizes MPC hyperparameters online. | 1. Stability during parameter switching<br>2. Sim-to-Real gap | 1. Heavy-duty autonomous trucking<br>2. High-performance racing |
| Certified Safety Verification | **Parallel Architecture:** Performance planner + Deterministic safety filter/validator. | 1. Feasibility in OOD states<br>2. Verifying semantic/VLM safety | Robotaxis in specific Operational Design Domains (ODDs) |

### 5.1. World Model-Centric MPC

The first promising direction involves integrating generative world models (Ha and Schmidhuber, 2018) as the internal prediction engine within the MPC framework. This approach represents an evolution of the DDOC paradigm: rather than relying on simplified kinematic equations or standard NN approximations, the controller utilizes a generative world model—such as Gaia-1 or ADriver-I—to perform state rollouts in a latent or video space. By functioning as a high-fidelity simulator within the optimization loop, this architecture enables the MPC to “imagine” diverse future outcomes in complex, interactive scenarios that traditional physics-based models fail to capture.

However, practical deployment faces significant challenges regarding closed-loop stability and real-time feasibility. A critical open question is ensuring long-horizon consistency, as generative models are prone to “hallucinations” or fidelity drift over time, which can mislead the optimizer into generating unsafe trajectories. Furthermore, the substantial computational latency of large generative models conflicts with the strict frequency requirements of vehicle control. Consequently, near-term applications will likely prioritize high-fidelity closed-loop simulators for offline validation or deployment in low-speed logistics and shuttle vehicles, where semantic reasoning capabilities take precedence over millisecond-level dynamic response.

### 5.2. Hybrid Learning for Customized Control

To reconcile the “black-box” nature of end-to-end learning with the rigor of control theory, this direction adopts a bi-level optimization framework. The upper-level functions as a “Driver Digital Twin”, employing IRL or LLMs to infer user intent from driving data or natural language instructions. This level outputs interpretable MPC parameters—specifically cost function weights and dynamic constraint boundaries—which are then fed into the lower level. The lower level remains a standard MPC solver, utilizing these parameters to generate trajectories that satisfy kinematic feasibility and safety constraints.

A critical unresolved challenge is balancing the conflict between personalization and safety. As discussed in Section 4.5 regarding constraint ambiguity, human drivers often treat traffic rules flexibly, whereas hard constraints in MPC can lead to infeasibility or overly conservative behavior. Integrating these fuzzy human preferences into rigorous safety constraints without compromising the system’s certified stability boundary presents a significant obstacle. Furthermore, achieving few-shot adaptation—tuning the controller to a new user with minimal data rather than requiring extensive training sets—remains a bottleneck for scalable commercial deployment. Given these challenges, the most immediate applications lie in L2+ Navigate on Autopilot and Highway Pilot systems. Unlike current commercial solutions that rely on static, manually selected modes, hybrid learning enables these systems to continuously tune following distances and lane-change aggressiveness. This evolution aims to align vehicle behavior with the user’s specific naturalistic driving style, thereby reducing user anxiety.

### 5.3. RL-Based Meta-Control Architecture

This direction advances the “System Self-Optimization” paradigm introduced in Section 4.4 by establishing a hierarchical RL-based meta-control architecture. Unlike standard RL which outputs direct control actions, the proposed architecture employs a RL agent as a high-level tuner that dynamically optimizes the hyperparameters of the low-level MPC—such as the prediction horizon, cost function weights, and terminal constraints—based on real-time environmental states and performance feedback. In this bi-level framework, the RL agent focuses on long-term strategic adaptation, while the MPC ensures high-frequency kinematic feasibility and instant safety.

However, realizing this self-evolving mechanism online presents critical open questions regarding closed-loop stability and computational tractability. A primary theoretical challenge is ensuring stability during the dynamic switching of control parameters. As the RL agent updates weights online, the system effectively becomes a switching system, where guaranteeing a common Lyapunov function across all configurations is non-trivial. Furthermore, the computational burden scales significantly when integrating an RL inference step with an iterative MPC solver, particularly when the optimization landscape is non-convex due to complex constraints. Bridging the "Sim-to-Real" gap also remains difficult, as an RL policy trained in simulation may fail to converge to optimal parameters under the stochastic disturbances of the real world. Given these characteristics, near-term applications will likely focus on heavy-duty autonomous trucking and high-performance racing. In trucking scenarios, vehicle dynamics change drastically with payload variations, making fixed-parameter MPC inefficient; a self-tuning controller can adaptively adjust model parameters and control gains to maintain stability without manual recalibration. Similarly, in autonomous racing or limit-handling scenarios, where tire friction coefficients and vehicle responsiveness evolve rapidly, this architecture allows the system to push the vehicle's handling envelope by dynamically relaxing or tightening constraint boundaries in real-time.

### 5.4. Certified Safety Verification

To reconcile the uncertainty of learning-based models with strict safety requirements, this strategy advocates for an architecture where an independent safety verification module operates alongside the learning-based planner. In this setup, the learning model generates candidate actions to maximize performance, while the verification module acts as a deterministic filter enforcing safety constraints. For instance, the Predictive Safety Filter (He et al., 2025) or the safety validator (Zhang et al., 2025a) exemplify this approach. Similarly, Huang et al. (Huang et al., 2026) utilized a lightweight risk field model to provide runtime safety guarantees.

A critical open question is establishing theoretical completeness for data-driven constraints. While filters like those proposed by He et al. (He et al., 2025) ensure constraints are met based on historical data, proving recursive feasibility when the vehicle encounters out-of-distribution states—where historical data provides no "safe" precedent—remains mathematically challenging. Additionally, verifying high-dimensional semantic safety is a major hurdle. Unlike control constraints, validating the cognitive safety of a VLM's decision requires bridging the gap between vague natural language safety rules and deterministic logic verification, and ensuring that the validator itself does not suffer from misinterpretation or hallucinations (Zhang et al., 2025a). Given the rigorous nature of this architecture, near-term applications will prioritize robotaxis operating in specific operational design domains. In these driverless scenarios, the architecture provides the necessary deterministic safety layer to handle corner cases that models might mishandle, satisfying the certification requirements for commercial operation where human intervention is impossible.

## 6. Conclusions

This paper presents the first systematic survey of the Data-Driven Optimal Control (DDOC) paradigm within the context of autonomous driving motion planning. Addressing the limitations of traditional hierarchical architectures in adapting to complex scenarios, as well as the 'black-box' opacity and lack of safety guarantees inherent in end-to-end AI methods, we first propose a pioneering evolutionary roadmap for DDOC-based motion planning. Furthermore, we demonstrate the paradigm's exceptional capabilities across three critical dimensions essential for next-generation autonomous driving: customization, dynamic adaptation, and self-tuning. Based on this analysis, we identify pivotal future research directions. Ultimately, this roadmap seeks to transcend functional feasibility, steering autonomous driving systems toward a future of safe, trustworthy, and efficient commercial deployment.

## Acknowledgments

This paper is partially supported by National Science and Technology Major Project (No. 2022ZD0115500), National Natural Science Foundation of China (Grant No. 52302412 and 52372317), Yangtze River Delta Science and Technology Innovation Joint Force (No. 2023CSJGG0800), Shanghai Automotive Industry Science and Technology Development Foundation (No. 2404), Xiaomi Young Talents Program, the Fundamental Research Funds for the Central Universities (22120230311), Tongji Zhongte Chair Professor Foundation (No. 000000375-2018082), and Shanghai Sailing Program (No. 23YF1449600).